\documentclass[10pt,twocolumn,letterpaper]{article}

\usepackage[pagenumbers]{wacv} 

\definecolor{wacvblue}{rgb}{0.21,0.49,0.74}
\usepackage[pagebackref,breaklinks,colorlinks,allcolors=wacvblue]{hyperref}
\usepackage[most]{tcolorbox}
\usepackage{xcolor}
\usepackage{subcaption}
\usepackage{graphicx}

\def\wacvPaperID{1542} 
\def\confName{WACV}
\def\confYear{2026}

\title{Unsupervised Memorability Modeling from Tip-of-the-Tongue Retrieval Queries}

\author{
\textbf{Sree Bhattacharyya}\textsuperscript{1} \quad
\textbf{Yaman K. Singla}\textsuperscript{2} \quad
\textbf{Sudhir Yarram}\textsuperscript{2} \\
\textbf{Somesh Singh}\textsuperscript{2} \quad
\textbf{Harini SI}\textsuperscript{2} \quad
\textbf{James Z. Wang}\textsuperscript{1}
\\[8pt]
\textsuperscript{1}The Pennsylvania State University \quad
\textsuperscript{2}Adobe Media and Data Science Research
\\[6pt]
{\tt\small sfb6038@psu.edu, behavior-in-the-wild@googlegroups.com}
}

\begin{document}
\maketitle
\begin{abstract}
Visual content memorability has intrigued the scientific community for decades, with applications ranging widely, from understanding nuanced aspects of human memory to enhancing content design. A significant challenge in progressing the field lies in the expensive process of collecting memorability annotations from humans. This limits the diversity and scalability of datasets for modeling visual content memorability. Most existing datasets are limited to collecting aggregate memorability scores for visual content, not capturing the nuanced memorability signals present in natural, open-ended recall descriptions. In this work, we introduce the first large-scale unsupervised dataset designed explicitly for modeling visual memorability signals, containing over 82,000 videos, accompanied by descriptive recall data. We leverage tip-of-the-tongue (ToT) retrieval queries from online platforms such as Reddit. We demonstrate that our unsupervised dataset provides rich signals for two memorability-related tasks: recall generation and ToT retrieval. Large vision-language models fine-tuned on our dataset outperform state-of-the-art models such as GPT-4o in generating open-ended memorability descriptions for visual content. We also employ a contrastive training strategy to create the first model capable of performing multimodal ToT retrieval. Our dataset and models present a novel direction, facilitating progress in visual content memorability research. \footnote{Code available at: \url{https://github.com/sreebhattacharyya/web_scale_memorability}.}
\end{abstract}
    
\section{Introduction}
\label{sec:introduction}

\begin{figure*}[t]
    \centering
    \includegraphics[width=0.9\linewidth]{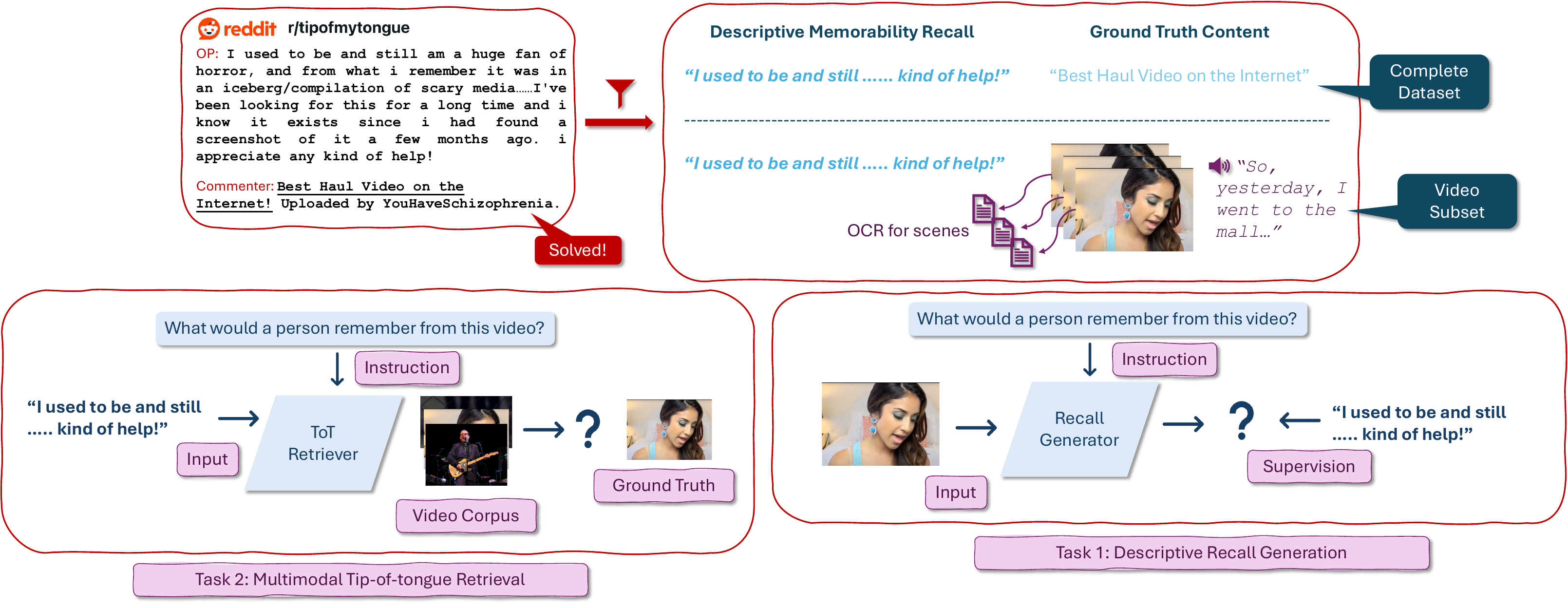}
    \caption{Our complete data collection and task pipeline. We use Tip-of-the-Tongue (ToT) search posts from Reddit (top left), and collect data through a rigorous filtering process. This leads us to obtain data points that are essentially recall-content pairs. The original Reddit search query becomes the descriptive recall, as it is what a user tries to identify the content from their memory recalls, and subsequently expresses on the platform. The correct content is retrieved from within the comments made to the posts. We also create a video-based subset of the data by downloading the raw visual information from YouTube and providing additional details such as audio transcripts and OCR. We propose two tasks using our dataset: Descriptive Recall Generation and Multimodal ToT retrieval. We also present \textsc{ToT2MeM-Recall} and \textsc{ToT2MeM-Retrieval}, respectively, to generate descriptive memorability recall or perform multimodal retrieval, trained on our dataset.}
    \label{fig:pipeline}
\end{figure*}

In today's digital age, the constant stream of content delivered through the internet and social media inundates people with more information than ever before. Our minds, however, retain only part of such information effectively. Several factors influence what persists in memory, including characteristics of the information itself, as well as individual differences. Studying content memorability, especially for visual media, has long been a research goal within psychology, cognitive science, and computer science. Creating systems that can automatically model content memorability has diverse applications, including in education \cite{cowan2014working}, advertising, marketing \cite{friestad1986emotion}, and design \cite{venni2021aesthetics, owendesigning}. Within the field of Artificial Intelligence (AI), applications include creating personalized agents \cite{park2023generative}, tailored recommender \cite{beheshti2020towards}, and retrieval systems \cite{roediger2013retrieval, schwartz2011tip}, or providing a better computational understanding of human intelligence \cite{khosla2015understanding}. Past research within AI has mainly focused on understanding and modeling visual content memorability, spanning the modalities of images and videos. Adding to the findings from human studies in psychology \cite{brady2008visual, konkle2010conceptual}, these computational efforts have found that a large part of memorability can be attributed to factors intrinsic to the visual content \cite{isola2011makes, khosla2015understanding, harini2025long}, such as semantic concepts, emotions, or object categories. This is crucial to understanding and incorporating memorable properties within content, which can potentially generalize for all individuals interacting with it. \\

\indent While several past research works study visual memorability, they usually depend on obtaining human annotations to create a memorability dataset. The most common method for collecting such data, especially for short-term recall, is in the form of a visual memory game \cite{isola2011makes}. For experiments studying long-term recall, the process becomes further challenging, as individual participants are required to first view the visual content, and then respond to memorability-related questions a significant amount of time later. In \cite{harini2025long}, for example, the authors state that the total time for curating the dataset (which contains only 2205 samples) was two years. This makes the data curation process less scalable, both in terms of the number of data points collected and the detail of memorability signals obtained from the participants. For example, most works express memorability for visual content through \textit{singular scores}, aggregated over the data collected from all participants. This does not account for the rich signals that can be leveraged from \textbf{open-ended recall descriptions} of visual content, which can provide directions on exactly \textbf{what} is memorable, beyond knowing only \textbf{how} memorable something is (recognition). \\
\indent In our work, we address this gap by harnessing web-scale data from online Tip-of-the-Tongue (ToT) \cite{schwartz2011tip} search forums such as Reddit \footnote{For example, the Tip Of My Tongue thread on Reddit: https://www.reddit.com/r/tipofmytongue/}. These online ToT search forums operate on the underlying principal of ``wisdom of crowds" \cite{galton1907vox, surowiecki2005wisdom}: users who are trying to remember some content they have interacted with in the past, but are unable to produce precise identifiers to recall or retrieve them, post inexact descriptions on the forums, while other participants respond with their best guesses for the potential answer. The ``correct answer'' is finally chosen by the question poser. The advantage of data from such platforms is the availability of descriptive memory recall signals - in the form of the search query posted - as well as the opportunity to collect diverse, large-scale data of free-form interactions. This helps us scale to multiple content domains, such as videos, books, movies, and songs, and long-term descriptive recall for modeling memorability. Additionally, using the ``correct answers" present in the comments of such posts, we construct a large-scale video-based subset of our data, which we use to model recall generation and perform ToT retrieval. 

Our main contributions can be summarized as: 

\begin{itemize}
    \item We introduce \textsc{\textbf{ToT2MeM}}, the first large-scale dataset for modeling visual content memorability with \textbf{open-ended recall signals} (Section \ref{sec:dataset}). Our dataset consists of over 470,000 content-recall pairs, spanning multiple content domains and genres. Further, we present a video-based subset of our data, \textsc{\textbf{ToT2MeM-Video}}, containing 82,500 video-recall pairs, and the raw visual content (eg., scenes), and audio. 
    \item We provide a detailed analysis of our large-scale dataset, presenting insights on how pervasive content features, like genre, impact memorability (Section \ref{sec:data_analysis}). 
    \item Using \textsc{ToT2MeM-Video}, we propose the novel task of descriptive recall generation. We present a trained model \textsc{ToT2Mem-Recall}, capable of generating human-like recall signals for video content. We show that this model generalizes to high-quality out-of-distribution human data for descriptive recall generation and a memorability-related ranking task (Section \ref{sec:recall}). 
    \item We also create \textsc{ToT2MeM-Retrieval}, using a contrastive training strategy \cite{jiang2025vlm2vec} with our dataset, and present the first solution for multimodal ToT retrieval (Section \ref{sec:tot_retrieval}). 
\end{itemize}
We visualize our entire data and task framework in Fig. \ref{fig:pipeline}. We plan to release all of our data and models publicly.



\section{Related Work}
\label{sec:related_work}

\begin{table*}
    \centering
    \resizebox{\linewidth}{!}{
    \begin{tabular}{cccccc}
    \toprule
    Dataset & \# Samples & Memorability Signal & Modality & Domain / Type & Data Collection Setting \\
    \midrule
    LaMem & 60,000 & Recognition & Images & Open / General & Competition-based Experiment \\
    SUN & 2,222 & Recognition & Images & Open / General & Competition-based Experiment \\
    MemCat & 10,000 & Recognition & Images & Open / General & Competition-based Experiment \\
    Memento10k & 10,000 & Recognition & Videos & Single-action amateur videos & Competition-based Experiment \\ 
    MediaEval & 1500 & Recognition & Videos & Short clips from FlickR and Twitter & Competition-based Experiment \\
    VideoMem & 10,000 & Recognition & Videos & Single-action staged video & Competition-based Experiment\\ 
    LAMBDA & 2,205 & Recognition and Recall & Videos & Advertisement & Natural Experiment \\ 
    \midrule
    \textsc{\textbf{ToT2MeM}} & \textbf{470,000 (total), 82,500 (videos)} & \textbf{Recall} & \textbf{Text, Images, and Videos} & \textbf{General Entertainment} & \textbf{Unsupervised in-the-wild interactions} \\
    \bottomrule
    \end{tabular}}
    \caption{Comparison between current and popular memorability datasets and our dataset.}
    \label{tab:comparison}
\end{table*}

\subsection{Visual Content Memorability}
\label{sec:visual_content_memorability}

A majority of early research efforts in visual content memorability focus on images \cite{khosla2015understanding, isola2011makes, goetschalckx2019memcat, borkin2016beyond, khosla2012memorability, khosla2012image, borkin2016beyond}, with more recent works venturing into studying memorability for video content \cite{cohendet2019videomem, newman2020multimodal, kiziltepe2021annotated, harini2025long}. 
Most of these studies model "recognizability" using an aggregate memorability score. Only two past research efforts \cite{harini2025long, borkin2016beyond} collect recall-related information. \cite{harini2025long} collects descriptive recall signals from participants, between 24 - 72 hours after they have interacted with the videos. However, the specific recall content is not used for modeling memorability. \cite{borkin2016beyond}, on the other hand, collects recall descriptions from participants, in the presence of a blurred version of the original images in the study. Both of these studies also focus on specific domains - advertisements and visualizations, respectively. Thus, descriptive recall signals have remained relatively underexplored in past works. Further, most of the past studies have also focused on small-scale human data collection, often in a competitive, in-the-lab setting. We compare the scale of our dataset with other popularly used memorability-related datasets in Table \ref{tab:comparison}.

\subsection{Tip-of-the-Tongue Retrieval}
\label{sec:tot_retrieval}

Tip-of-the-Tongue (ToT) Retrieval is a subset of general known-item search \cite{lee2006known}, where the user searches for an item they have interacted with in the past, while being in a "tip-of-the-tongue" state \cite{schwartz2011tip}.
Several past works study ToT retrieval by building relevant datasets and benchmarks \cite{arguello2021tip, jorgensen2020kinda, bhargav2023music}. Almost all of the existing methods have focused on retrieval in a text-to-text setting, often reducing multimodal content to a textual format (eg, plot descriptions as a proxy for movies) \cite{arguello2021tip}. Most existing datasets are also domain-specific, and relatively limited in scale, while some expand to several casual search domains (eg., books, movie, games, videos, etc.) for qualitative \cite{meier2021towards} and quantitative analysis \cite{frobe2023large} of the query content and quality.
Several works also present baseline methods for ToT retrieval, including standard methods like BM25 to LLM-based re-ranking of outputs \cite{borges2024generalizable}. However, throughout existing literature, the challenging nature of the task has been repeatedly highlighted. 
\cite{bhargav2023music} show that neither individual parts of a ToT query nor LLM-based reformulations yield good retrieval performance, while \cite{meier2021towards} find that pre-retrieval indicators cannot predict how long it takes for a query to be solved. Building on these challenges, we introduce the first large-scale dataset and method for multimodal ToT retrieval, demonstrating that LLM-based embeddings can substantially improve performance.

\section{\textsc{ToT2MeM}: Using Tip-of-Tongue Retrieval Queries for Descriptive Memorability}
\label{sec:dataset}

In this section, we describe our procedure to collect data and present a detailed analysis of the same.


\paragraph{Data Filtering and Collection from Reddit:}
\label{sec:data_collection}
To collect large-scale memorability data, we focus on threads where Reddit users post descriptions about diverse content items (eg., movie, videos, images, books, quotes, etc.), usually from their memory of having interacted with the item in the past, in a ``tip-of-the-tongue" description fashion \cite{schwartz2011tip}, and pose a question, wanting to know what the exact content item was. Other Reddit users then respond in the comments, usually providing their best guesses of what they think the item could be. We collect data from 2431 unique Reddit threads known for hosting ToT-style searches between January 2017 and June 2022. 

We exclude posts that contain NSFW markers or are created by bots. 
The total number of posts we find after such initial filtering are 1,979,167. We then remove posts that are marked as deleted (about 2\% of the total). 
Out of the remaining posts, 471751 posts are marked as ``solved", meaning that some user was able to post the correct reference to the exact content item being searched. 
We validate the ``solved" status of posts through multiple checks. Firstly, we consider whether an appropriate (Solved) tag appears in the flair CSS class for the post, as is common in many Reddit threads. Then, we iterate through each comment made on the post and identify whether the post author replies explicitly to any of the comments, confirming that the comment provides the correct answer. Further, we also check whether the moderator bot for a given subreddit thread confirms the same correct answer within the post comments. 
Finally, we verify the ``correct answer" obtained through these mechanisms by passing the entire post, along with all comments, to a small LLM (DeepSeek-R1 Distilled on LLaMA 8B) \cite{guo2025deepseek} in a few-shot setting, where it independently identifies the correct response by scanning the entire text content of a post. In Appendix \ref{app:data_collection}, we provide further details of the validation process and additional data statistics. 

We focus specifically on solved posts, as they show that the recall descriptions provided in the original post were \textit{reliable enough} to find the correct answer. Our dataset comprises all 470,000 solved posts, spanning multiple domains of content items.

\begin{figure*}[htbp]
    \centering
    \begin{subfigure}[b]{0.31\textwidth}
        \includegraphics[width=\linewidth]{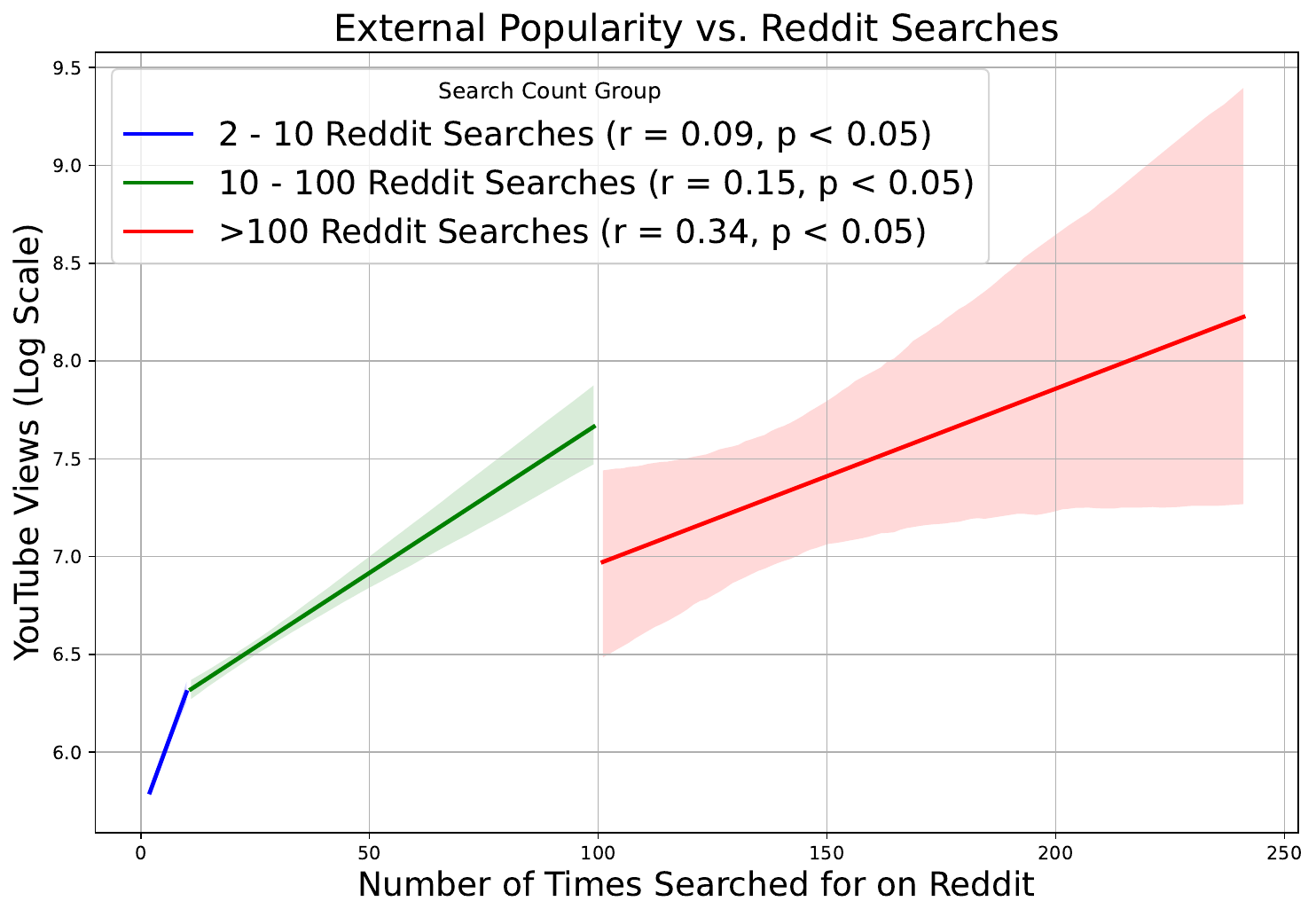}
        \caption{}
        \label{fig:yt-vs-search}
    \end{subfigure}
    \hfill
    \begin{subfigure}[b]{0.31\textwidth}
        \includegraphics[width=\linewidth]{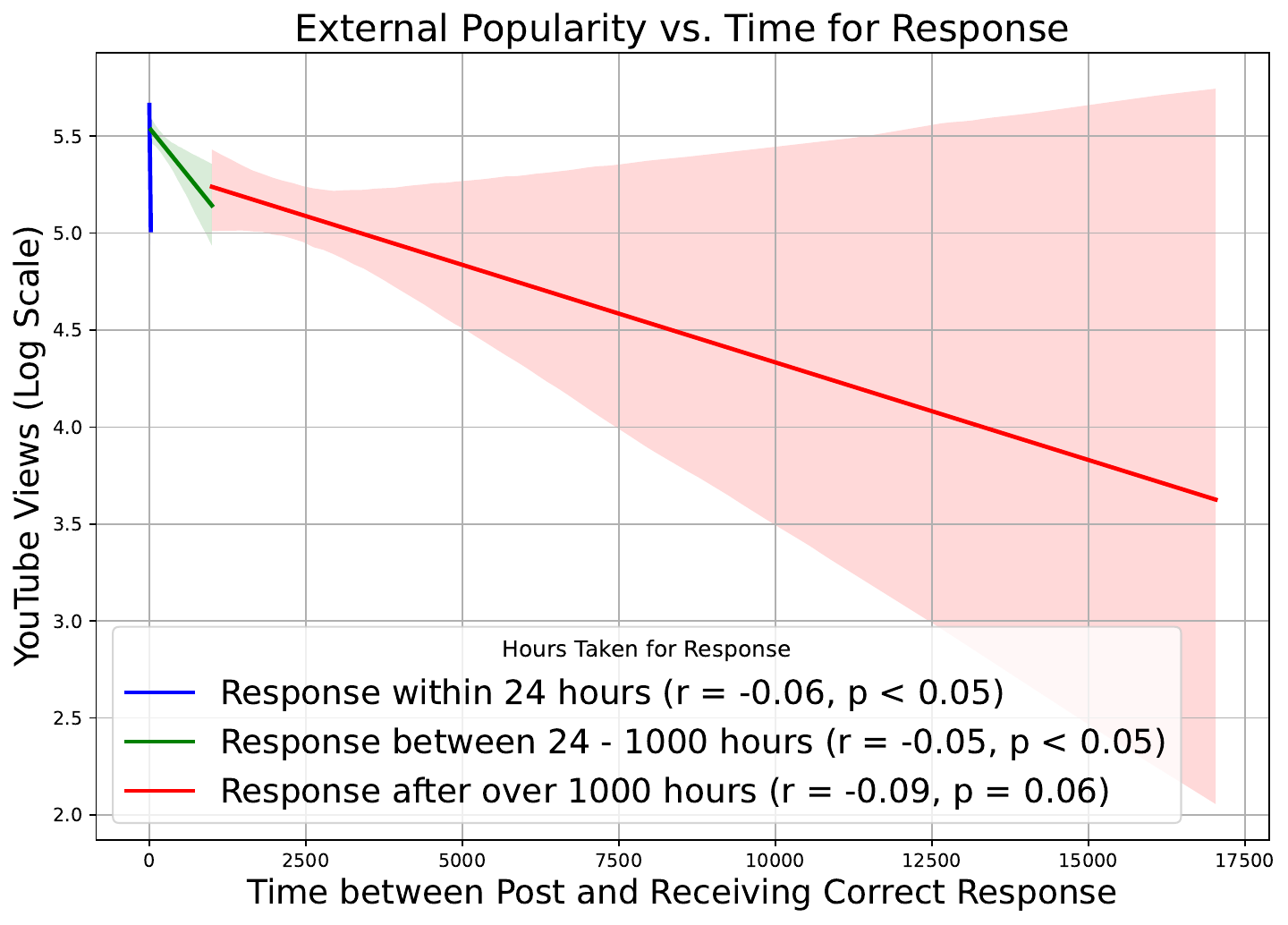}
        \caption{}
        \label{fig:yt-versus-response}
    \end{subfigure}
    \hfill
    \begin{subfigure}[b]{0.31\textwidth}
        \includegraphics[width=\linewidth]{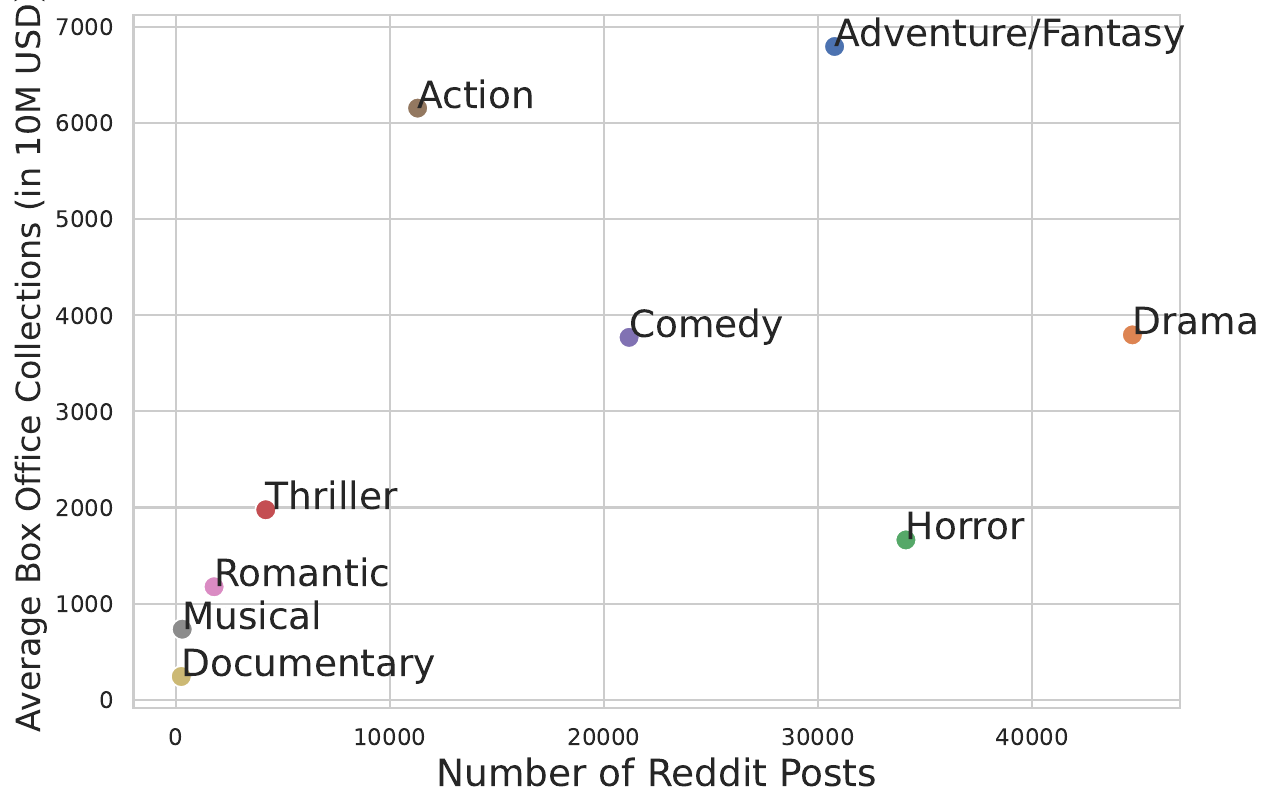}
        \caption{}
        \label{fig:box-office-versus-posts}
    \end{subfigure}
    \caption{(a) Correlation between external popularity (measured using YouTube views) and number of times searched on Reddit. Zoomed-in graphs for each group (based on the search count) are included in the Appendix \ref{app:data_analysis}. We also include the analysis using Wikipedia page views as a measure of popularity in the Appendix. (b) Correlation between external popularity (YouTube views) of the content and the time between the original post about the content is made, and when someone comments with the correct answer. (c) Relationship between popularity, as measured by the average box office collections, of different genres, and the number of posts found in ToT forums. A more detailed picture of the relationship between genre popularity and searches is presented in Appendix \ref{app:data_analysis}.}
    \label{fig:popularity-analysis}
\end{figure*}

\begin{figure*}[t!]
    \centering
    \begin{subfigure}[b]{0.40\textwidth}
        \includegraphics[width=\linewidth]{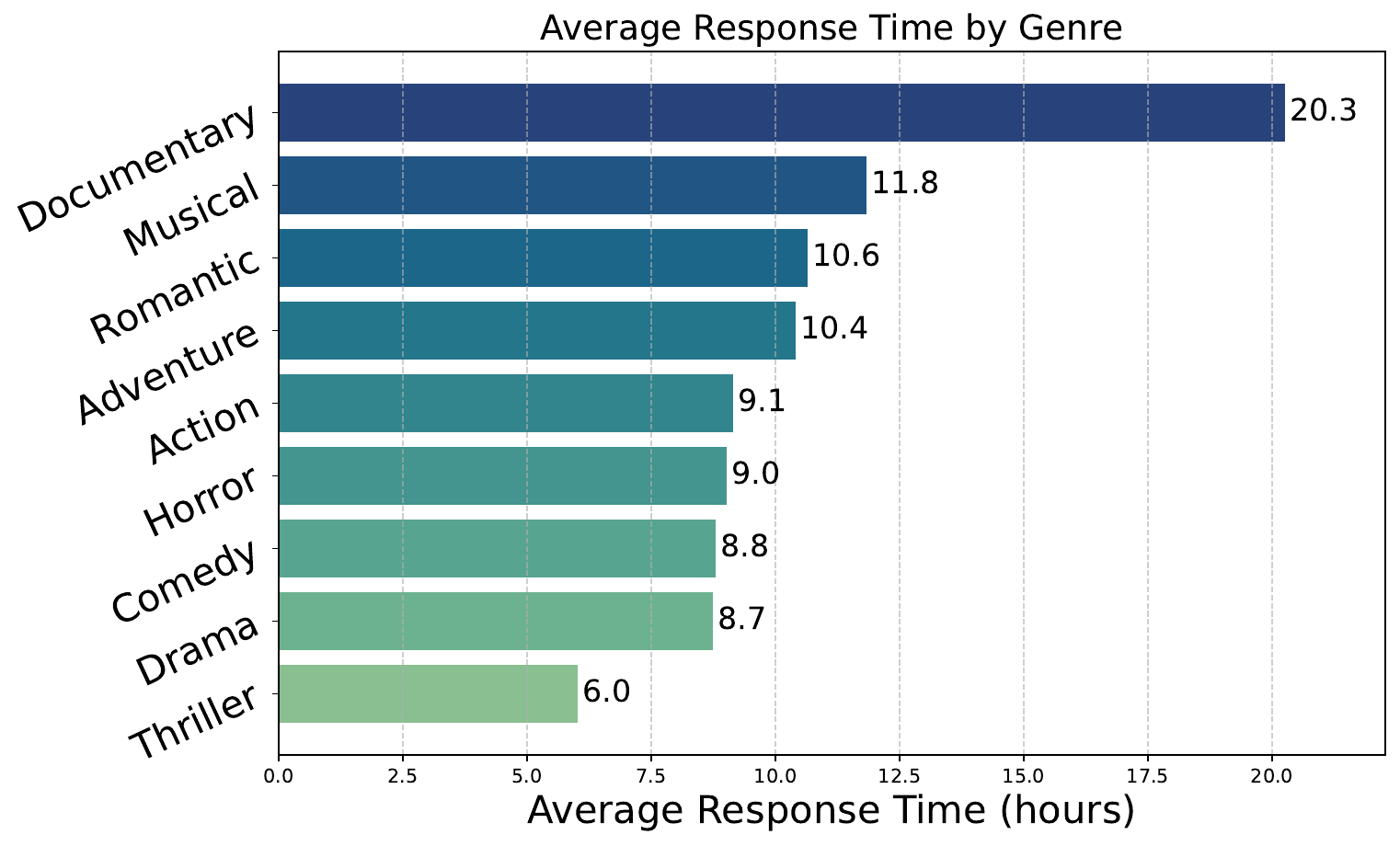}
        \caption{}
        \label{fig:response-time-by-genre}
    \end{subfigure}
    \hfill
    \begin{subfigure}[b]{0.52\textwidth}
        \includegraphics[width=\linewidth]{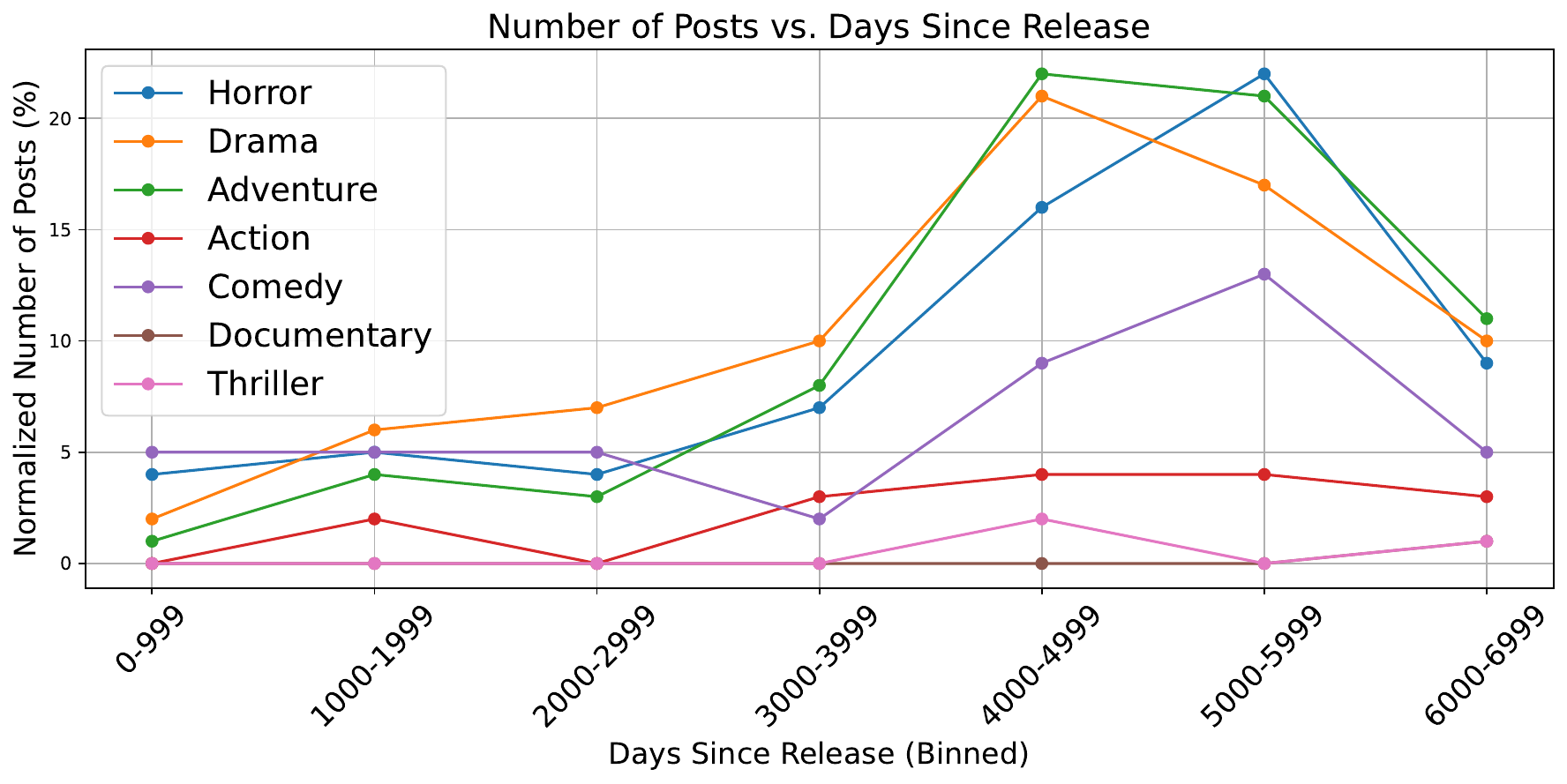}
        \caption{}
        \label{fig:number-posts-since-release}
    \end{subfigure}
    \caption{(a) Average Response Time (in Hours) for content belonging to each genre. Response time refers to the time elapsed between a post/search being made and the time when the correct answer is provided in the comments. (b) Comparison of searches made for content in each genre, with time (in days) since the release of that content (as obtained from the Wikipedia page creation date).}
    \label{fig:genre-analysis}
\end{figure*}

\paragraph{Creation of the Video-Based Data Subset \textsc{ToT2MeM-Video}:}
\label{sec:video_subset}

We find that YouTube videos are one of the most frequently searched content types on Reddit platforms (Fig. \ref{fig:top-content-distribution} in Appendix \ref{app:data_analysis}). Leveraging this, we also create a large-scale video dataset, presenting the raw visual information of videos, to be directly used for training models for downstream tasks. We begin by extracting all YouTube links embedded within posts. We discard posts where the search query does not contain a substantial text description (e.g., consists of only a video link in the query itself, or any media in any other modality, such as audio or image). Then, we filter out posts for which the correct answer is a video hosted in a domain other than YouTube. 

We further filter posts to remove private or deleted videos, and only retain videos that are less than 600 seconds in length. As an additional verification mechanism, we utilize DeepSeek-R1 LlaMa 8B \cite{guo2025deepseek} in an LLM-as-a-judge setting to obtain the name of the correct video from each post, while also cross-validating with metadata obtained for each video from YouTube. We also manually verify about 100 such responses to ensure that they match and do not find any errors. Following these stages of filtering, we are left with 82,500 recall-video pairs that form \textsc{ToT2MeM-Video}. 

For each video in the dataset, we then sample scenes using a context-aware adaptive scene detector \footnote{\href{https://www.scenedetect.com/docs/latest/api/detectors.html\#scenedetect.detectors.adaptive_detector.AdaptiveDetector}{AdaptiveDetector - SceneDetect}}.
We also detect texts displayed in the sampled scenes through automatic OCR \cite{cui2025paddleocr} and generate audio transcripts for videos using Whisper \cite{radford2023robust}. Then, using the OCR texts for each scene, we create a de-duplicated set of scenes for each video, including only consecutive scenes where the OCR changes. This helps reduce both the computational complexity of processing the visual information and contextual noise in the visual data. For additional details on the filtering process, see Appendix \ref{app:video_subset}. \\

\noindent \textbf{Connecting \textsc{ToT2MeM} with External Content Factors:}
\label{sec:data_analysis} 
\noindent Several past works have shown the influence of intrinsic content factors on memorability, such as emotions displayed, popularity \cite{harini2025long}, or the presence of certain objects \cite{newman2020multimodal}. We complement this with an analysis of more pervasive content-related factors, such as genre popularity. We present our findings as follows: \\ 
\noindent \textbf{(a) Popular Items Are Searched More Often and Retrieved Faster:}
We use views on YouTube (for videos, movies, trailers, songs, etc.) as a proxy for external popularity, and find (Fig. \ref{fig:yt-vs-search}), that external popularity is positively correlated with popularity on Reddit search forums. 
Although externally popular content items (such as a YouTube video watched many times) can be expected to remain stored in memory (because it is being watched so frequently), they may appear in ToT searches because users from different cultural or linguistic backgrounds might lack the exact vocabulary to find it directly.
\indent On the contrary, we find a weak negative correlation between external popularity and the time required for users on Reddit to come up with the correct answers to searches (Fig. \ref{fig:yt-versus-response}). Further, we also compare genre-wise popularity (using the average box office collections of each genre in North America across 2017-2022), with the number of searches made about posts belonging to each genre, in Fig. \ref{fig:box-office-versus-posts}. We find that although Action and Adventure are popular in terms of revenue, Drama is the genre searched for the most. \\
\noindent \textbf{(b) Genre-Specific Analysis}: We then take a deeper look at genre-specific properties of the posts. Firstly, we study how quickly, on average, posts from each genre get solved. As demonstrated in Fig. \ref{fig:response-time-by-genre}, searches for content belonging to all popular genres get solved quickly, while searches for more niche content, such as documentaries, take the longest time. Second, we study how the number of searches about content within a specific genre change, with respect to time elapsed since its initial release. Most posts on ToT forums are created a long time after the original content is released, with most genres seeing a spike around 8 - 10 years after release. Content belonging especially to popular genres like Adventure, Drama, or Horror is particularly remembered (or attempted to be remembered), even long after it was created. 

\noindent To explore whether our unsupervised dataset (in particular \textsc{ToT2MeM-Video}) can suffice for modeling content memorability, we propose two novel tasks:
\begin{itemize}
    \item \textbf{Descriptive Recall Generation}, where we train a model using our dataset, to generate human-like memorability signals in the form of open-ended text.
    \item \textbf{Tip-of-the-Tongue (ToT) Retrieval}, where we explore whether the functionality served by online ToT forums (like Reddit) can be replicated automatically, through a model trained using our data.
\end{itemize}

\section{Task 1: Descriptive Recall Generation}
\label{sec:recall}

\begin{table*}
    \centering
    \begin{tabular}{ccccccc}
    \toprule
    Model & BLEU & METEOR & ROUGE-1 & ROUGE-2 & ROUGE-L & BERTScore \\
    \midrule
    GPT-4o (zero-shot) & 0.193 & \underline{0.197} & 0.204 & 0.027 & 0.117 & 0.81 \\
    InternVL-2.5 8B (zero-shot) & \underline{0.211} & 0.196 & \underline{0.216} & 0.059 & \underline{0.148} & \underline{0.82} \\
    InternVideo-2.5 8B (zero-shot) & 0.189 & 0.159 & 0.196 & 0.033 & 0.128 & \underline{0.82} \\
    Qwen-2 VL 7B (zero-shot) & 0.173 & 0.192 & 0.024 & \underline{0.123} & 0.122 & \underline{0.82} \\
    Qwen-2.5 VL 7B (zero-shot) & 0.196 & 0.153 & 0.2 & 0.021 & 0.126 & \underline{0.82} \\
    Qwen-2.5 Omni 7B (zero-shot) & 0.083 & 0.173 & 0.12 & 0.024 & 0.072 & \underline{0.82} \\
    \midrule
    \textsc{ToT2MeM-Recall} & \textbf{0.242} & \textbf{0.293} & \textbf{0.304} & \textbf{0.152} & \textbf{0.251} & \textbf{0.85}\\
    
    \bottomrule
    \end{tabular}
    \caption{Primary results for evaluation on the descriptive recall generation task. The top scores are shown in bold, while the second-highest scores are underlined.}
    \label{tab:recall_1}
\end{table*}

\begin{table*}
    \centering
    \begin{tabular}{ccccccc}
    \toprule
    Model & BLEU & METEOR & ROUGE-1 & ROUGE-2 & ROUGE-L & BERTScore\\
    \midrule
    GPT-4o (zero-shot) & 0.16 & 0.20 & 0.177 & 0.029 & 0.105 & 0.823 \\
    InternVL 2.5 8B (zero-shot) & 0.14 & \underline{0.19} & 0.16 & 0.03 & 0.11 & 0.82 \\ 
    Qwen 2 VL 7B (zero-shot) & \underline{0.21} & 0.17 & \underline{0.22} & \underline{0.04} & \underline{0.13} & \underline{0.84} \\
    Qwen 2.5 VL 7B (zero-shot) & 0.19 & 0.18 & 0.2 & 0.03 & 0.12 & 0.83 \\
    Qwen 2.5 Omni 7B (zero-shot) & 0.04 & 0.12 & 0.07 & 0.02 & 0.05 & 0.79 \\
    \midrule
    \textsc{ToT2MeM-Recall} (zero-shot) & \textbf{0.27} & \textbf{0.37} & \textbf{0.29} & \textbf{0.14} & \textbf{0.24} & \textbf{0.86} \\
    \bottomrule
    \end{tabular}
    \caption{Results on the LAMBDA Recall Dataset \cite{harini2025long}. The topmost scores are shown in bold, and the second-highest scores are underlined.}
    \label{tab:recall_1_lambda}
\end{table*}

The first task we propose to benchmark our dataset on, deals with predicting or generating, in a human-like manner, descriptive recall statements for visual content. In this section, we describe the proposed task setting, the experimental setup, and results for recall generation. 
\\

\noindent \textbf{Task Setup.} As described in Section \ref{sec:dataset}, our video-based data subset contains the following: the raw visual content (scene images), the corresponding OCR for each scene, and an audio transcript for the entire video. With this, there also exists the ground truth descriptive recall text, obtained from the original Reddit query \footnote{For training, we include all sentences of the original recall signals except ones that describe purely episodic memory to mitigate hallucinated responses about episodic memory.}. 
The goal of the recall generation task then is: given any visual media \(V\) and the associated information in the form of input, and a task instruction \(I\) (constant across all input samples), a given model \(M(\theta)\) should predict a sequence \(S'\) that describes what a person might remember from the given visual content, using the original ground truth recall \(S\) as the source of supervision. Thus, we simply have, \(M(V,I;\theta) = S\). We consider a pre-trained Vision-language Model (VLM) as our initial model \(M(\theta)\). \\


\noindent \textbf{Training and Evaluation.} We fine-tune the QwenVL 2.5 7B model \cite{bai2025qwen2}, in a parameter-efficient manner, using LoRA \cite{hu2022lora}, on \textsc{ToT2MeM-Video}. To show the effectiveness of the use of our dataset, we compare our results with various state-of-the-art multimodal baselines, capable of dealing with multi-image or video inputs, evaluating them in a zero-shot manner. Precisely, among open-source models, we evaluate InternVL 2.5 8B \cite{chen2024internvl}, QwenVL 2.5 7B \cite{bai2025qwen2}, and Qwen Omni 2.5 7B \cite{xu2025qwenomni} with multi-image inputs in a zero-shot manner, while also comparing with InternVideo-2.5 8B \cite{wang2025internvideo} to which we provide direct video inputs. Among proprietary models, we evaluate GPT4-o \cite{hurst2024gpt}. The baselines are chosen primarily owing to their superior performance in general multimodal and video understanding tasks \cite{hendrycks2020measuring, li2024mvbench}. Zero-shot evaluation is chosen for the baselines, as the inputs already contain multiple images; providing few-shot examples becomes further confusing (especially for open-source models) and expensive. We utilize popular text generation metrics from the literature, such as BLEU \cite{papineni2002bleu} and ROUGE \cite{lin2004rouge}, and also include METEOR \cite{banerjee2005meteor} and BERTScore \cite{devlin2019bert} to capture semantic similarity. \\

\noindent \textbf{Results.} The primary results are presented in Table \ref{tab:recall_1}. Our model \textit{outperforms the compared baselines on all of the metrics}, demonstrating that our dataset helps capture the nuances of open-ended memorability descriptions. Importantly, our model is based on an unmodified Qwen2.5-VL-7B backbone and differs from the baselines only in being fine-tuned on our dataset. This setup ensures that the comparison directly reflects the value of our dataset rather than architectural differences \footnote{We show, similarly, the effectiveness of using our dataset to tune another strong baseline, InternVL 8B, in Appendix \ref{app:recall_generation}.}. It also demonstrates how general multimodal reasoning ability and world knowledge, which models like GPT4-o excel at, might not be sufficient for predicting recall signals, given its unintuitive nature \cite{khosla2015understanding}. We also assess the statistical significance of the improvement of \textsc{ToT2MeM-Recall} on the top-performing baseline (InternVL 2.5 8B) using Wilcoxon's test \cite{wilcoxon1992individual}, and find that the improvements are highly statistically significant (\(p << 0.05\)). For additional experiments, including ablations, see Appendix \ref{app:recall_generation}. We also provide qualitative examples and failure analysis in Appendix \ref{app:qualitative_examples}.

\noindent \textbf{Zero-shot Transfer.} To understand whether the performance gains simply stem from having a distributional advantage, we further evaluate the baselines and our model on an unseen dataset: LAMBDA \cite{harini2025long}. LAMBDA is the only other known descriptive recall dataset, containing a total of 2205 samples. The high-quality recall signals in this dataset are collected from human participants through a supervised lab experiment. We show that our model, despite being trained on our unsupervised dataset, generalizes zero-shot and again outperforms all other baselines (Table \ref{tab:recall_1_lambda}). This implies that the memorability signals captured from \textsc{ToT2MeM-Video} are general, relevant, and aligned with high-quality recall signals provided by humans.


\begin{table}[t!]
    \centering
    \begin{tabular}{cc}
    \toprule
         Model & Accuracy (\%) \\
         \midrule
         InternVL-2.5 8B & 18.3 \\
         Qwen 2 VL 7B & 79.28 \\
         \textsc{ToT2MeM-Recall} & 82.04 \\
         \bottomrule
    \end{tabular}
    \caption{The ranking@1 accuracy of our model and two other strong baselines on the Prompt Recall Ranking Task. All models are evaluated zero-shot.}
    \label{tab:ranking_evaluation}
\end{table}

\noindent \textbf{Human Evaluation.} To further validate the robustness of our dataset, we examine whether \textsc{ToT2MeM-Recall} can generalize to tasks beyond its training setup using high-quality human data collected in a supervised setting. Building on the recognition task studied in \cite{harini2025long}, we introduce a \textit{Prompt Recall Ranking (PRR) task}. We conduct a human study with 163 participants from an academic institution. Participants were shown the original videos from the LAMBDA dataset, and after a 48-hour delay, were instructed to recreate the most memorable scene from a single advertisement video using a generative model (Stable Diffusion v2.1) \cite{rombach2022high}. Each participant provided natural-language prompts corresponding to their memory of the advertisement, iteratively refining up to five prompt variations per scene, resulting in a total of 815 prompts. Additional details about the data collection process are presented in Appendix \ref{app:human_evaluation}. This procedure yielded a dataset linking (a) the original advertisement videos from LAMBDA \cite{harini2025long}\footnote{Note that the original LAMBDA dataset is also curated in a supervised lab setting.}, (b) human-generated prompts for reconstructing memorable scenes (referred to as diffusion prompts), and (c) the final images generated from these prompts using Stable Diffusion. 

In the PRR task, the model is given an original video along with five human-generated diffusion prompts—only one of which is correct—and must identify the correct corresponding prompt. Results for this evaluation are shown in Table \ref{tab:ranking_evaluation}. We find that \textsc{ToT2MeM-Recall}, although trained solely on our unsupervised dataset and for the distinct task of descriptive recall generation, generalizes well to the PRR task and significantly outperforms strong zero-shot baselines such as InternVL and Qwen 2 VL \footnote{For an additional discussion on the performance gap between the baselines, see Appendix \ref{app:human_evaluation}}. This demonstrates that training on \textsc{ToT2MeM} helps models acquire generalizable memorability signals that transfer effectively to human-curated data, even in a distinct, previously unseen task setting.

\section{Task 2: Tip-of-the-Tongue Retrieval}
\label{sec:tot_retrieval}

\begin{table*}
    \centering
    \resizebox{\linewidth}{!}{
    \begin{tabular}{cccccc}
    \toprule
    Embedding Model & Document Set & Effective Task & Recall/MRR{@}1 
    & {@}10 & 
    {@}100 \\
    \midrule
    BGE & Query Set & T2T Retrieval & 96.5/96.5 
    & 98.7/97.4 
    & 99.6/97.4 \\
    \midrule
    BGE & Generated Recall by InternVL & T2T Retrieval & 6.6/6.6 
    & 17.1/9.4 
    & 37.1/10.1 \\
    BGE & Generated Recall by \textsc{ToT2MeM-Recall} & T2T Retrieval & 9.1/9.1 
    & 21.2/12.6 
    & 43/13.3 \\ 
    \midrule
    GME-QwenVL & Original Videos & T2V Retrieval & 5.7/5.7 
    & 17.6/9 
    & 41.6/9.8 \\
    GME-QwenVL w/ Instruction & Original Videos & T2V Retrieval & 7.1/ 7.1 
    & 20.4/10.9 
    & 45.5/11.8 \\
    VLM2Vec-V2.0 & Original Videos & T2V Retrieval & \underline{10.02/10.02} 
    & \underline{25.4/14.5} 
    & \underline{48.5/15.3} \\ 
    \midrule
    \textsc{ToT2MeM-Retrieval} & Original Videos & T2V Retrieval & \textbf{10.9/10.9} 
    & \textbf{27.5/15.7} 
    & \textbf{50.6/16.5}\\
    \bottomrule
    \end{tabular}}
    \caption{Tip-of-the-tongue retrieval evaluation. The topmost performance is shown in bold, and the second-highest scores are underlined. T2T retrieval stands for Text-to-Text Retrieval, and T2V retrieval stands for Text-to-Video Retrieval.}
    \label{tab:tot_retrieval}
\end{table*}

Tip-of-the-tongue (ToT) retrieval differs from standard retrieval tasks in that the search queries are often inexact, vague, or even misleading. Using our dataset, we introduce the novel task of \textit{multimodal ToT retrieval} and present a model designed for this purpose, \textsc{ToT2MeM-Retrieval}.

\noindent \textbf{Task Setup.} Similar to traditional retrieval settings, the task involves searching within a set of documents \(D = \{d_1, d_2, ..., d_n\}\) to find the most relevant document \(d_i\) for a given query \(q\). In our case, \(q\) is the noisy textual description provided by a user, the document corpus is a set of videos, and the target document is the correct video the user is attempting to recall. To enable retrieval, both queries and documents must be mapped into a shared embedding space. We achieve this by adapting a vision–language model (VLM) and using its last-layer hidden state representations.

Formally, each training instance consists of a query–video pair \((q, t^+)\), where \(t^+\) is the ground-truth video, and a set of hard negative videos \(t^-\) mined based on semantic similarity (details in the Supplementary). Following \cite{jiang2025vlm2vec}, we prepend a task-specific instruction to the queries. Each training query input \((q)\) thus combines: (a) the task instruction, (b) the video descriptions (transcripts and OCR text), and (c) sampled scene images from the video. The training target is the associated recall description \((t)\). Note that this training configuration reverses the evaluation setting, where recall text is used as the query and videos are the documents.

We obtain last-layer representations \((h_q, h_{t^+})\) for \((q, t^+)\) and \((h_{t^-})\) for negatives \(t^-\), and optimize the standard InfoNCE loss \cite{jiang2025vlm2vec}:
\[
\min \mathcal{L} = - \log \left( 
\frac{ \phi(h_q^{\text{inst}}, h_{t^+}) }{ 
\phi(h_q^{\text{inst}}, h_{t^+}) + \sum_{t^- \in \mathcal{N}} \phi(h_q^{\text{inst}}, h_{t^-}) } 
\right),
\]
where \(\mathcal{N}\) includes both hard negatives and in-batch negatives. After training, the adapted embedding model is used to represent both the queries and the document corpus in the same space for retrieval.

\noindent \textbf{Evaluation Setup.} We evaluate several baselines for the retrieval task to compare with our model. Across all setups, the query is fixed as the ground-truth ToT search description provided by the user, while the baselines vary in how they represent the document corpus. Below, we describe each setting and the models used:
\begin{itemize}
    \item \textbf{Ground Truth Recall as the Document Set} (Topline): We explore what the best possible performance is for the ToT retrieval task. For this, we use the \textit{ground truth recall statements} themselves as the document corpus to serve as a proxy for the respective videos. In other words, the set of queries \(Q = \{q_1, q_2, .... , q_n\}\), and the document corpus, \(D = \{d_1, d_2, .... , d_n\}\) are \textit{identical}. Here, the retrieval task reduces to a text-to-text retrieval setting. As both of the queries and documents are in the text format here, we use BGE Embeddings \cite{chen2024bge} to embed the queries and documents into the same space. 
    \item \textbf{Generated Recall as the Document Set}: Similar to the topline, in this setting, we use \textit{recall statements generated by models} as the document corpus, where each generated recall statement would serve as a proxy for the respective videos. We use descriptive recall signals for each video in our dataset, generated in a zero-shot manner by InternVL 2.5 8B \cite{chen2024internvl}, as well as \textsc{ToT2MeM-Recall} from Section \ref{sec:recall}. In this setting, too, the task is reduced to text-to-text retrieval. Similar to the topline setting, we also embed the query (ground truth recall statements) and documents (generated recall statements), both of which are text-based, using BGE \cite{chen2024bge} to then use the embeddings for similarity-based search and retrieval. 
    \item \textbf{Original Videos as the Document Set}: This setting represents the original text-to-video retrieval task. We compare with several baseline embedding models, such as GME \cite{zhang2024gme}, and VLM2Vec-V2 \cite{jiang2025vlm2vec}. For our contrastive training strategy, we use VLM2Vec-V2 as the initial model, fine-tuning the unmodified architecture with \textsc{ToT2MeM-Video}. All of these embedding models are used simply to embed both queries (text recall statements) and documents (all videos) into the same embedding space, followed by a similarity matching-based retrieval process. 
\end{itemize}
For all of these settings, we report results on standard metrics used in retrieval - Recall@k and MRR@k. 

\noindent \textbf{Results.} Our results in Table \ref{tab:tot_retrieval} highlight both the difficulty of the task and the value of contrastive training on our dataset. First, the topline setting achieves only near-perfect scores, even though queries and documents are identical, showing how semantically similar user ToT queries can be.

\indent Second, we find that recall statements generated by our model (Section \ref{sec:recall}) serve as strong proxies for the corresponding videos. Retrieval performance using our generated recalls surpasses that of other state-of-the-art generators, and remains comparable to embedding-based methods such as GME that leverage the full video. This indicates that our recall signals effectively capture memorability-related information and can substitute for visual content.

Finally, our contrastively trained model \footnote{Built on the strongest zero-shot baseline, VLM2Vec, fine-tuned with our dataset.} outperforms all baselines across metrics. While its scores fall short of the topline, they are competitive in the broader ToT retrieval literature. Notably, even state-of-the-art methods for purely textual ToT retrieval -- a much simpler task -- achieve Recall@10 only in the 3–30\% range \cite{bhargav2023music, borges2024generalizable, arguello2021tip}. These results position our approach as an early but meaningful step toward robust multimodal ToT applications.

\section{Discussion and Conclusion}



\noindent \textbf{Data Quality, Noise, and Ethical Considerations.} We present the first large-scale dataset for modeling descriptive memorability signals, built from online tip-of-the-tongue (ToT) search platforms. To demonstrate its utility, we introduce two new tasks: descriptive recall generation and multimodal ToT retrieval. Training on this unsupervised web-scale dataset yields substantial gains over state-of-the-art baselines and enables fine-tuned models to generalize to two human datasets collected in controlled lab settings, including a novel zero-shot ranking task.

While its scale inevitably introduces minor noise - such as occasional errors in solved-answer detection, residual episodic signals, or OCR/ASR recognition errors - the strong empirical gains achieved on multiple benchmarks indicate that such noise does not hinder performance. This suggests the dataset provides high-quality supervisory signals for modeling long-term memorability. In future work, it could also support controlled studies on how large vision–language models handle noisy or ambiguous inputs, including whether such noise can induce hallucinations.

The dataset may also reflect the demographic and topical biases of Reddit, such as the dominant presence of entertainment-related content, searched for in English. To mitigate ethical risks, we release only post IDs, URLs, and derived features (frames, transcripts, OCR) without user identifiers, adhering to Reddit’s terms of service and fair-use principles. 

\noindent \textbf{Future Directions.} Our work opens up avenues for studying nuanced memorability signals and improved applications in practically relevant downstream tasks like tip-of-the-tongue retrieval. Several potential future directions exist for each of the tasks proposed. Using the detailed recall descriptions, interventions could be created for generating personalized and memorable content, providing an opportunity for applications in advertising. At the same time, exploration of ethical concerns with such applications remains critical to ensure users are not manipulated adversely for commercial gains. An empirical analysis of factors affecting memorability (e.g., emotions, episodic context, etc.) could also provide potential insights into the workings of human long-term memory. For multimodal ToT retrieval, advanced feature representations can be crafted or learned using memorability-specific modules. Scaling to the web level to ensure that multimodal ToT search can actually be performed, agentic search systems can also be used to solve the cold-start problem in retrieval. We hope that the public release of our dataset and methods can help pave the path for large-scale applications in the future.

{
    \small
    \bibliographystyle{ieeenat_fullname}
    \bibliography{main}
}

\clearpage

\appendix

\section{Data Collection From Reddit}
\label{app:data_collection}

In this section, we provide additional methodological details and results for our process of collecting data from Reddit. 

\subsection{Exploratory Data Statistics}

\begin{figure*}[htbp]
    \centering
    \begin{subfigure}[b]{0.24\textwidth}
        \includegraphics[width=\linewidth]{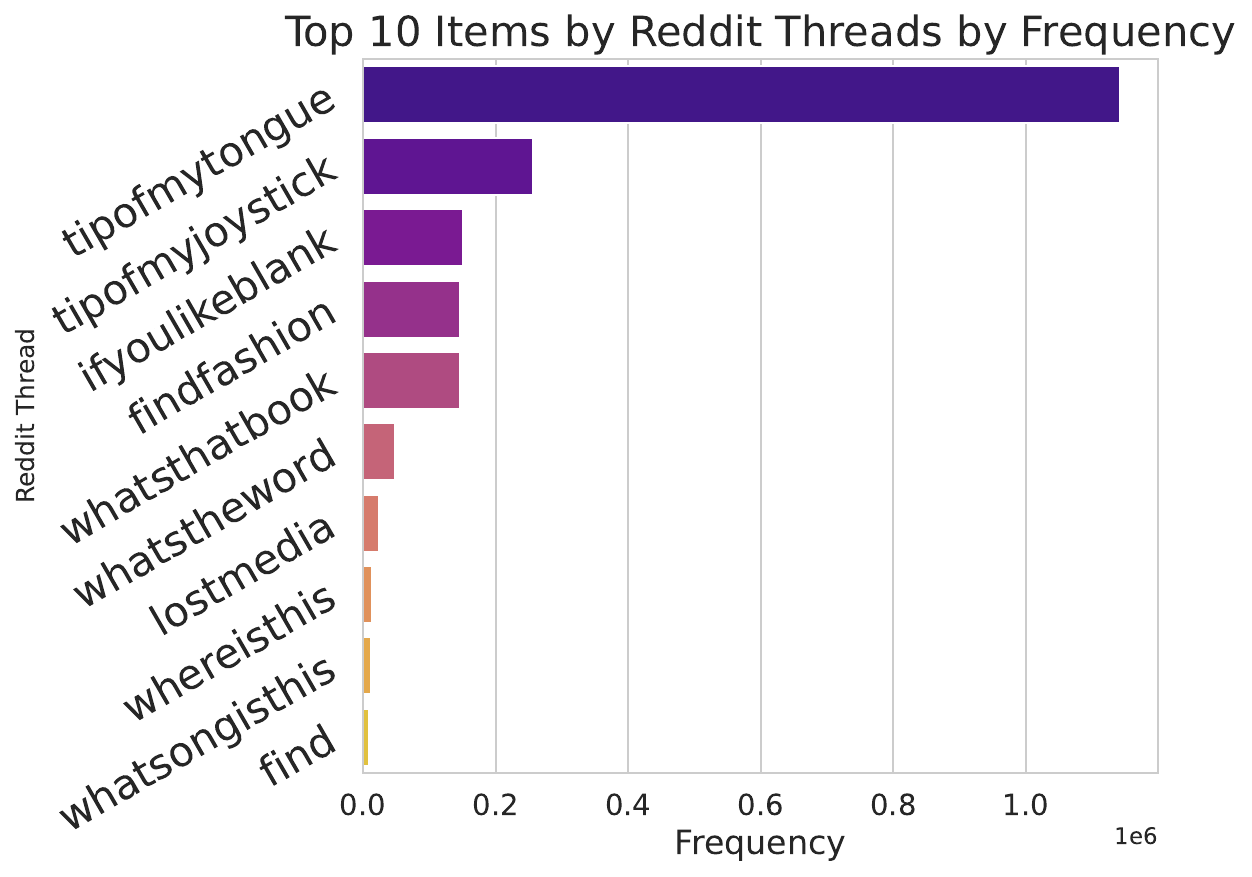}
        \caption{}
        \label{fig:top-threads}
    \end{subfigure}
    \hfill
    \begin{subfigure}[b]{0.25\textwidth}
        \includegraphics[width=\linewidth]{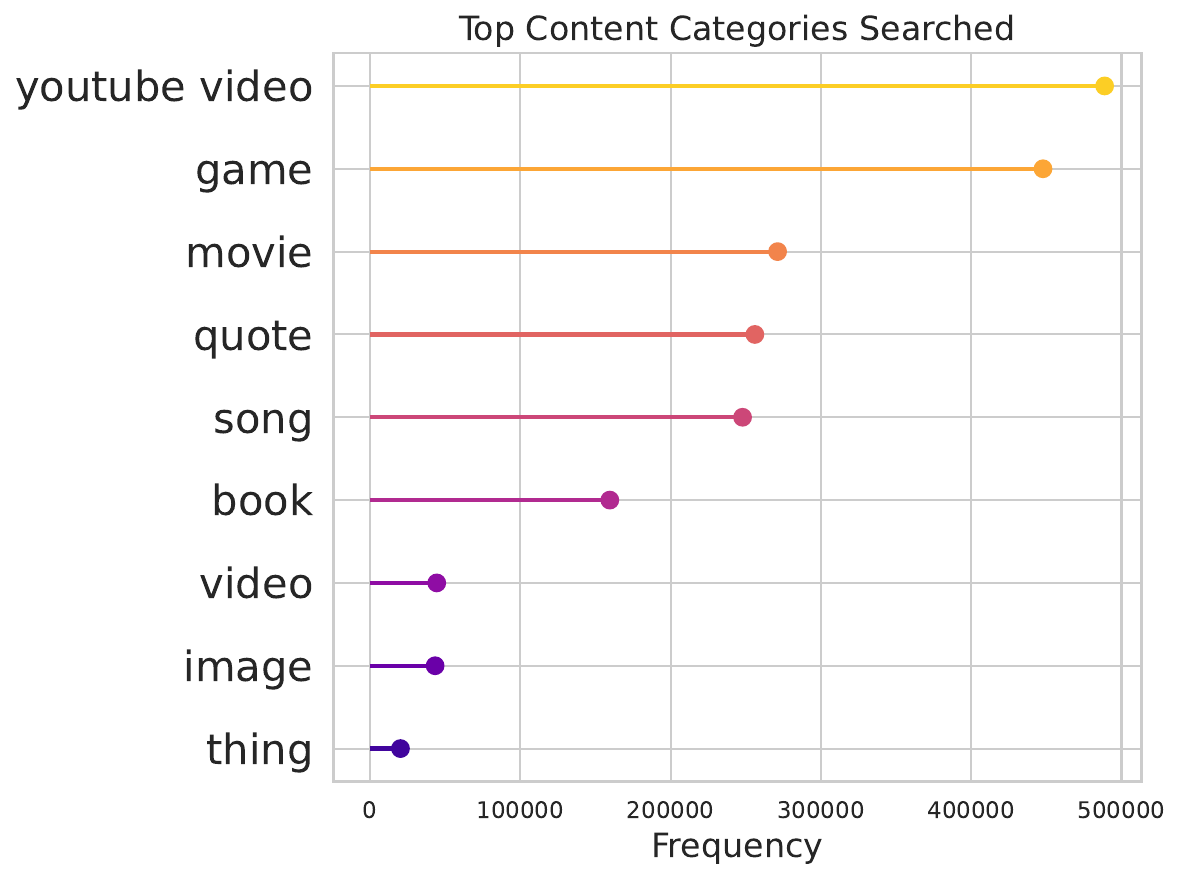}
        \caption{}
        \label{fig:top-content-distribution}
    \end{subfigure}
    \hfill
    \begin{subfigure}[b]{0.25\textwidth}
        \includegraphics[width=\linewidth]{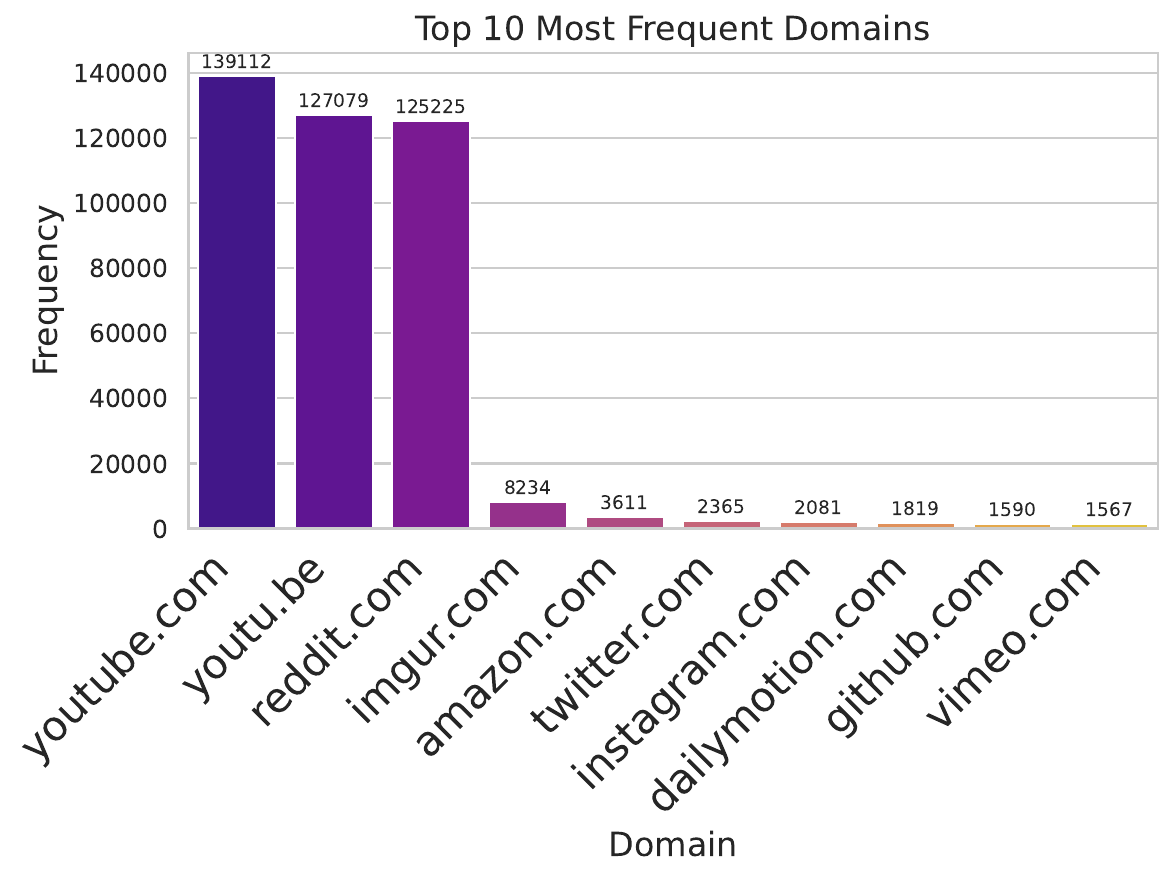}
        \caption{}
        \label{fig:top-domains}
    \end{subfigure}
    \hfill
    \begin{subfigure}[b]{0.24\textwidth}
        \includegraphics[width=\linewidth]{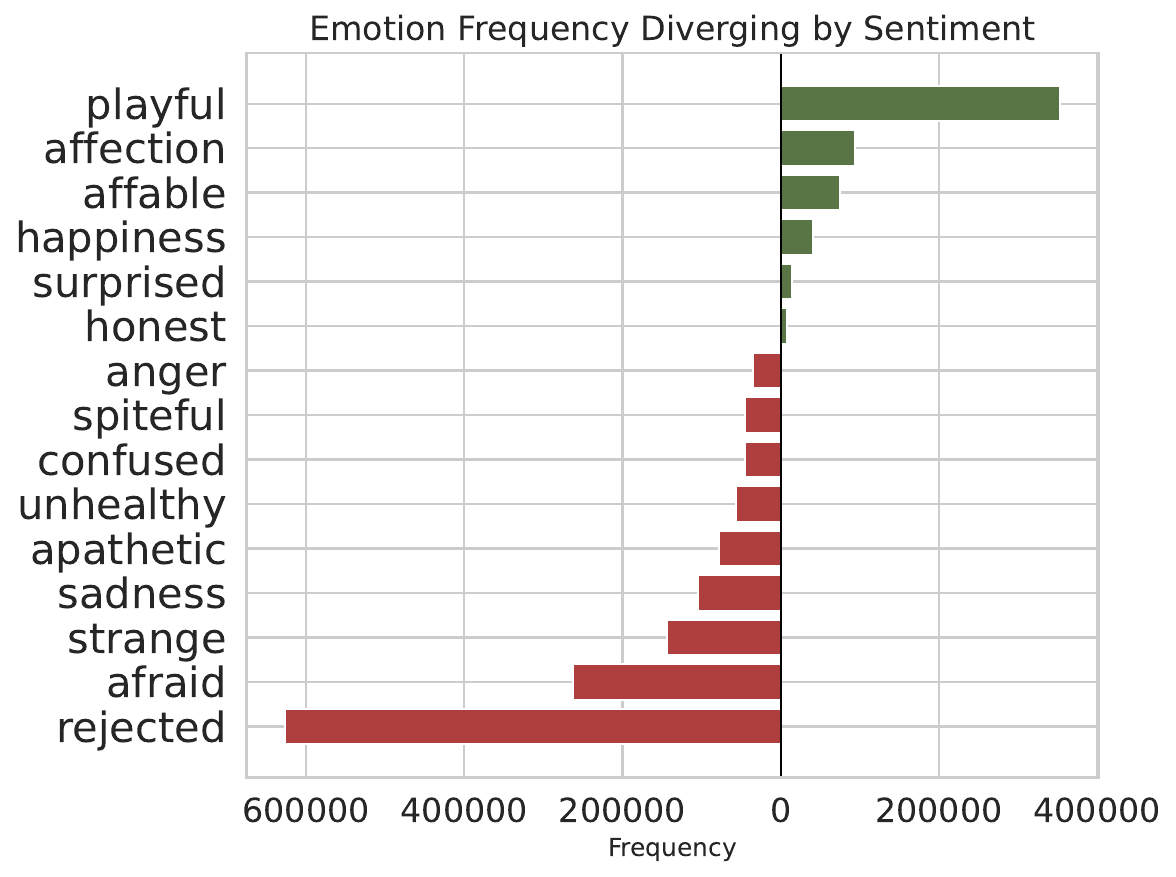}
        \caption{}
        \label{fig:top-emotions}
    \end{subfigure}
    \caption{(a): Top 10 Reddit threads used to construct our dataset. (b): Top 10 content types included in our dataset. (c): Top domains to which direct links are present in the dataset, indicating most usually that the correct content item is referred to using a link to these domains. (d): Distribution of emotions in the original Reddit posts. In other words, these are the emotions expressed within the recall signals. The analysis uses the culturally robust HICEM emotion model \cite{wortman2023hicem}.}
    \label{fig:metadata-info}
\end{figure*}

In this section, we provide some additional statistics of the entire \textsc{ToT2MeM} dataset collected from Reddit. Firstly, we show in Fig. \ref{fig:top-threads} the top-10 threads that are dataset is built upon. The top-10 threads contribute to over 90\% of the data points in the dataset. Then, in Fig. \ref{fig:top-content-distribution}, we show that some popular search items on these platforms include YouTube videos, games, and movies, which are usually all pertaining to casual or entertainment-related searches. Congruent with this, we also find a large presence of links from the YouTube domain in the comments, as shown in Fig. \ref{fig:top-domains}. We also empirically corroborate the finding that users in tip-of-the-tongue states (thereby, in these online communities) experience frustration more frequently \cite{elsweiler2007towards}, by showing that negative emotions are expressed in these search posts much more often than positive emotions (Fig. \ref{fig:top-emotions}).

\subsection{Validation of Correct Answers}

\begin{figure*}
    \centering
    \includegraphics[width=\linewidth]{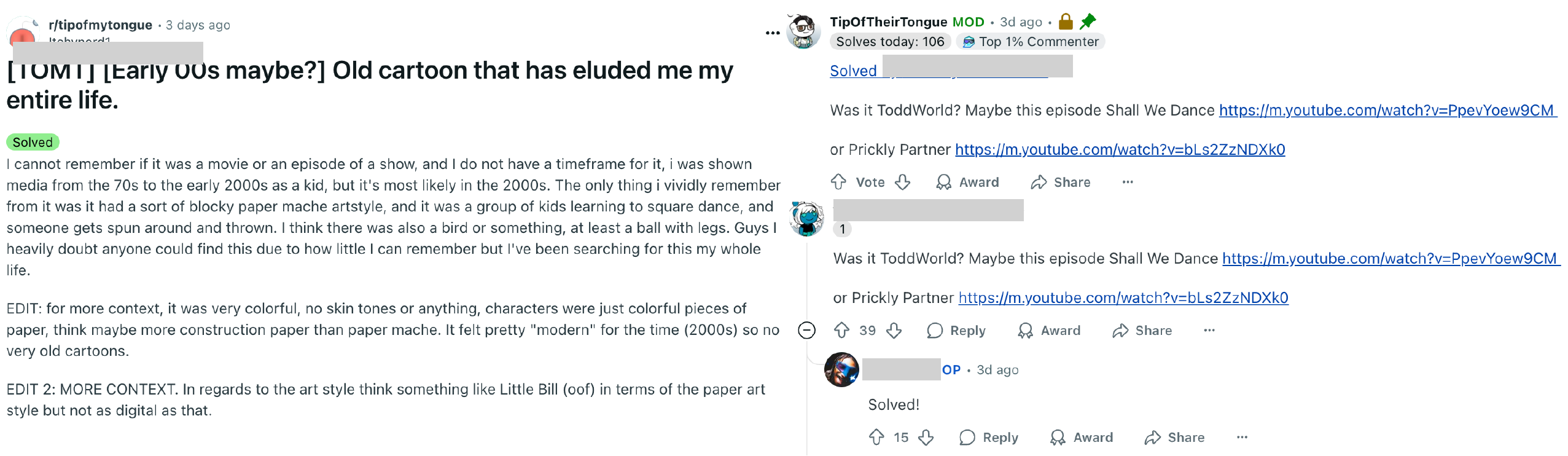}
    \caption{An example of different signals utilized for solving the valid answer. The thread moderator bot provides a comment, which is usually pinned, highlighting the correctly solved answer. Note that the thread moderator usually exactly copies the correct answer and provides it additionally. Further, the actual solving comment can also be found by tracking replies from the original poster, in case it provides directly confirming signals (such as in this case, by saying "Solved!").}
    \label{fig:valid_answer}
\end{figure*}

As mentioned in the Section \ref{sec:data_collection}, we use several parallel pipelines for validating the correct answer obtained from a post. Fig. \ref{fig:valid_answer} provides an example of a post from Reddit, where different identifiers of the valid answer can be found. The provided example also shows the case where two videos are linked as the ``correct answer". Using our automated resolution mechanism, we retain both correct answers in case multiple video links are attached to them, as it provides additional nuanced signals for recall. For example, if the same recall query could be related to multiple correct solutions, it could help in learning shared aspects or signals about memorability. Precisely, we use regular expressions to find the presence of links in the solved comments. If multiple links can be resolved, the correct answer is formatted as a list of links. However, if no links are found, and the correct answer consists of only text, the entire text is stored as the solved post. 

We also use LLM-based validation, where DeepSeek-R1 Distilled LLaMA 8B is used to resolve the name of the post. We provide the following prompt to the model, which also includes querying for other related information about the content: \\ 

\fbox{
  \parbox{0.9\linewidth}{
    \textbf{Basic Prompt Template:}
    "You will be given an online post, where a user asks about some content that they are trying to remember, but can only provide currently a vague description of, from their memory. You will also be provided with the response that is given to the user, which contains the name, or link of the correct content item that the user is trying to find about. Your task is, given this conversation, to respond clearly with the name (or link) and type of the content item that the user is asking about. You are required to answer strictly in a JSON format, as follows: {\"content name\": "name of the content item and the link to it, if present", \"category\": "could be movie, book, youtube video, game, song, quote, or anything", "genre": "could be anything among comedy, drama, horror, fantasy, adventure, or not applicable", "objects": "standard object categories", "emotions": "any common emotions that maybe present"}. Answer strictly in the JSON format, with the correct content item name, and category for the following user post and reply: ". 
    "Original Reddit post search query"
    "Correct Answer from the comments"
  }} 
\\

Through the LLM-based verification process, we find \(>\)95\% of the posts to have solved answers, coherent to what is found through the rule-based resolution process. Among the mismatched posts, we manually inspect 80 posts and find the response from the rule-based mechanism to be more reliable, as it includes the entire text portion of the solving comment. We thus retain the solution from the automated check as part of the dataset, instead of the resolution found by the LLM.

\section{Video-Based Data Subset}
\label{app:video_subset}

We collect and download all available YouTube videos linked in the comments of posts available in our dataset. Using this, we create a multimodal dataset of visual data-recall signal pairs,  We initially find around 120000 YouTube links embedded within the posts or comments. Among these initial YouTube videos, over 85\% are relatively shorter in length, under 10 minutes of duration. After filtering out links that are part of the post (the question or the search query), deleted posts, or deleted, gated, or corrupted media, our final dataset consists of 82,500 video-text pairs. The maximum number of de-duplicated scenes allowed for each video in the dataset is 30, to ensure that the entire input can fit into the context window of models, and limit computational complexity. All of the videos contain audio streams, and over 60\% of the videos contain at least one scene with a non-null associated OCR text.

\section{Connecting with External Factors}
\label{app:data_analysis}

To infer several subjective aspects of each post, such as the content category or genre it belongs to, or different objects, emotions present in it, we use DeepSeek Distilled LLaMA 8B \cite{guo2025deepseek}, in a few-shot manner. The same prompt, presented above, is used. 

Further, we use the APIs of the following websites to collect popularity data: Wikipedia, YouTube, iMDB, and the movie statistics platform The Numbers \footnote{https://www.the-numbers.com/}. 

We also study the memory content of the original posts (recall signals) in our dataset. We classify, in an unsupervised manner, each sentence within a Reddit post using DeepSeek-Distilled LLaMA 8B \cite{guo2025deepseek}. Sentences within a post may either describe content-related memory, describing, for example, a video an individual may have interacted with, episodic memory, providing additional context about such an interaction, or neither. Further, content-related memory sentences may describe semantic information about the content (eg., plot of a movie) or non-semantic information (eg., visual elements, location, release time, or actors in a movie). Using a few-shot approach, we tag sentences of each original post, finding that about 57\% of all sentences describe content-related memory, while 16\% describe personal, episodic memory, and the rest cannot be classified into either category. Within content-related information, the majority of sentences (68\%) describe non-semantic information, while the rest provide semantic descriptions of the content. 

\begin{figure*}
    \centering
    \includegraphics[width=\linewidth]{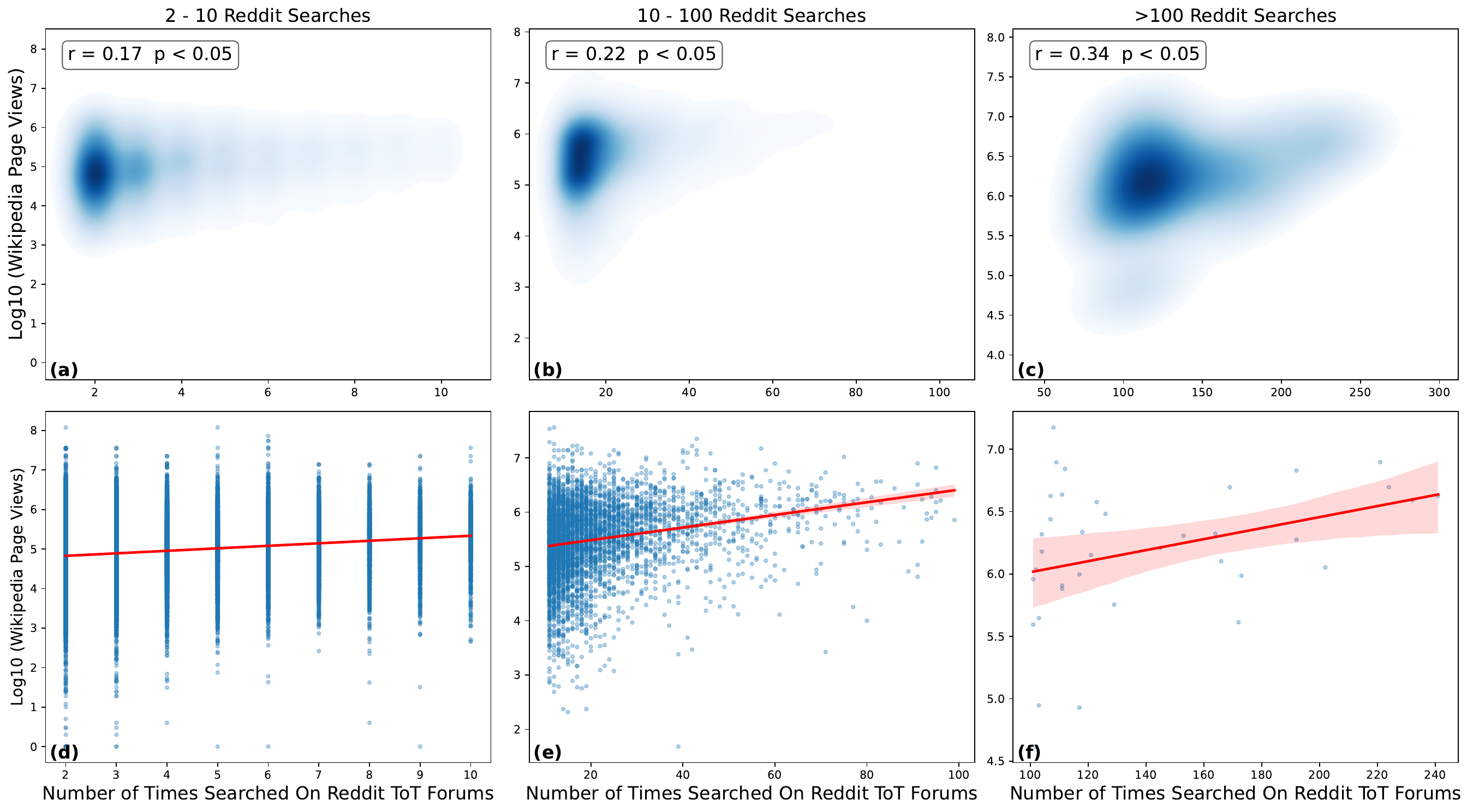}
    \caption{Correlation of Wikipedia-based popularity and number of searches on Reddit, for any given content item, shown in different groups, based on the number of searches made.}
    \label{fig:wiki-popularity-versus-reddit-search}
\end{figure*}

\begin{figure*}
    \centering
    \includegraphics[width=\linewidth]{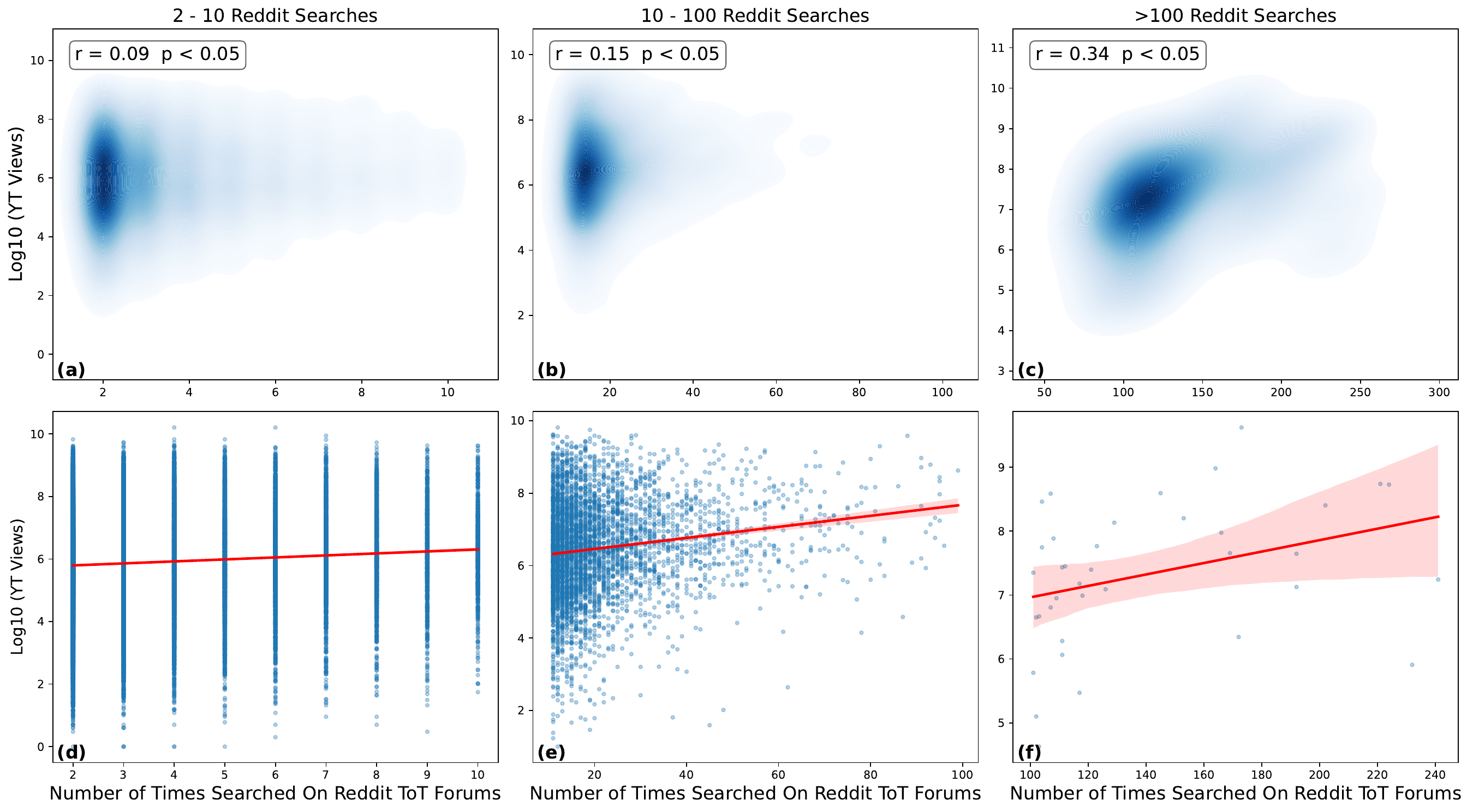}
    \caption{Correlation of YouTube-based popularity and number of searches on Reddit, for any given content item, shown in different groups, based on the number of searches made.}
    \label{fig:yt-popularity-versus-reddit-search}
\end{figure*}

\begin{figure*}
    \centering
    \includegraphics[width=\linewidth]{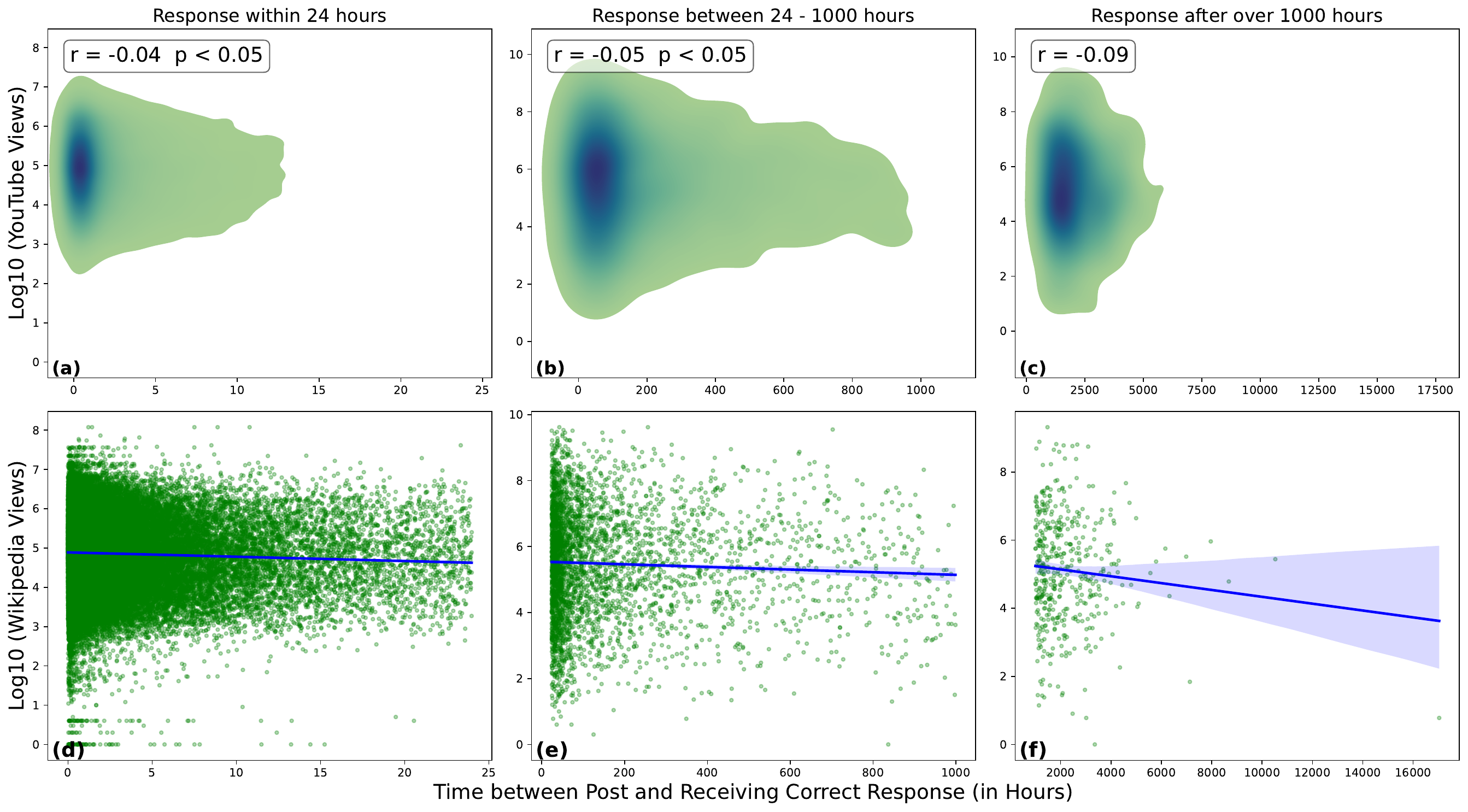}
    \caption{Zoomed-in correlation of Wikipedia-based popularity and response time on Reddit, to receive the correct answer.}
    \label{fig:wiki-popularity-versus-response-time}
\end{figure*}

\begin{figure*}
    \centering
    \includegraphics[width=\linewidth]{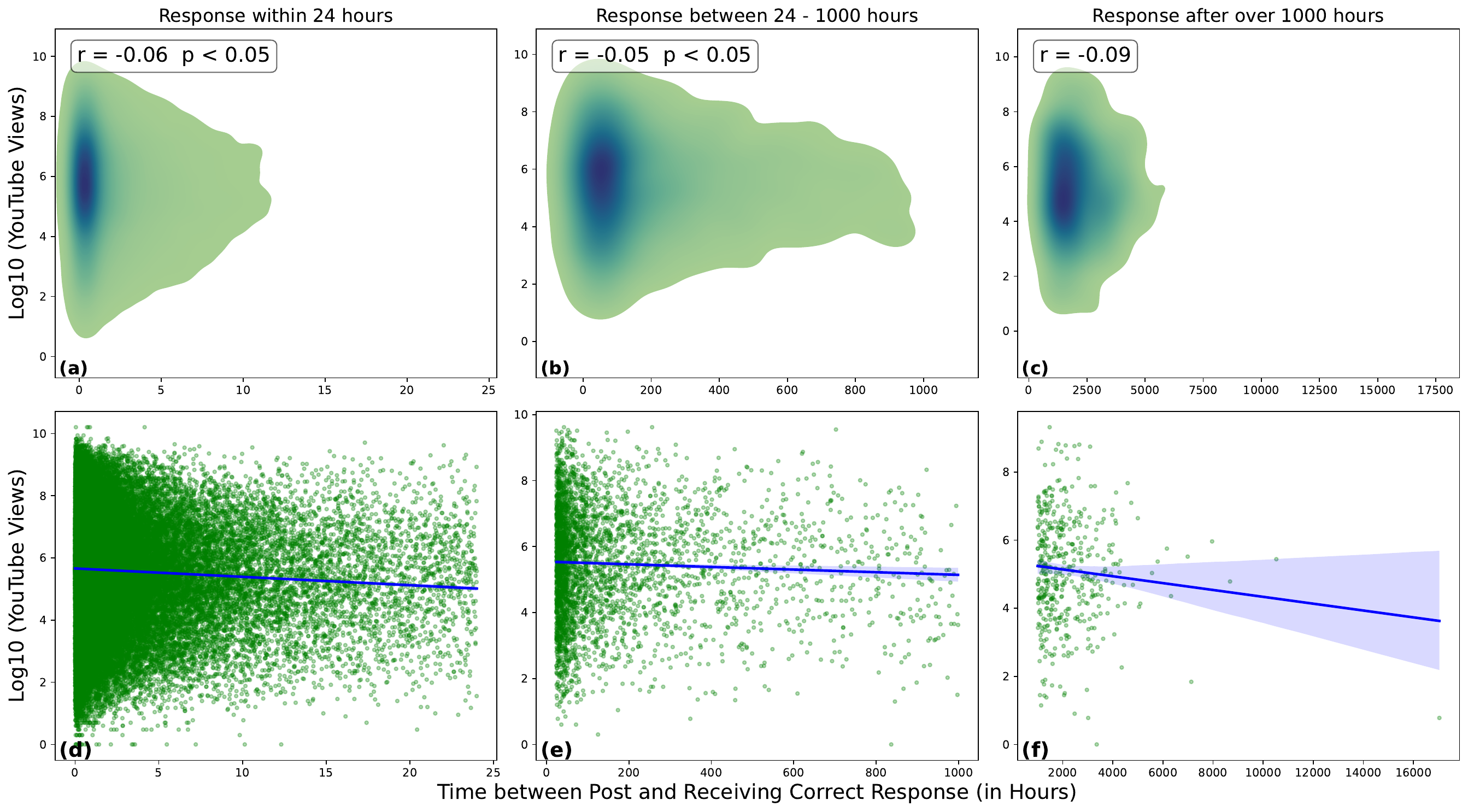}
    \caption{Zoomed-in correlation of YouTube-based popularity and response time on Reddit, to receive the correct answer.}
    \label{fig:yt-popularity-versus-response-time}
\end{figure*}

\begin{figure}[t]
    \centering
    \includegraphics[width=\linewidth]{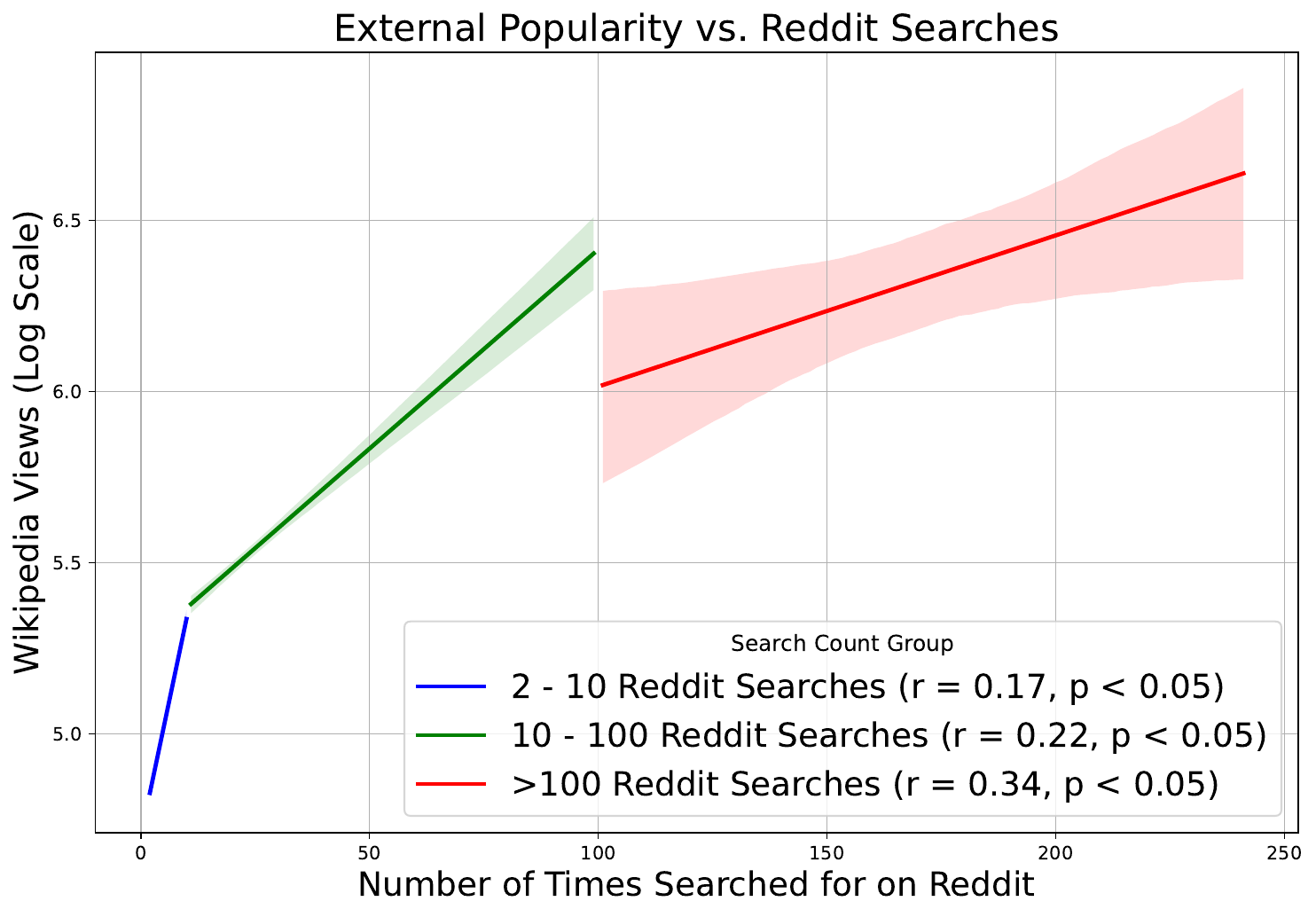}
    \caption{Condensed graph for correlation between Wikipedia page views (popularity) and number of searches on Reddit.}
    \label{fig:wiki-popularity-versus-reddit-condensed}
\end{figure}

\begin{figure}[t]
    \centering
    \includegraphics[width=\linewidth]{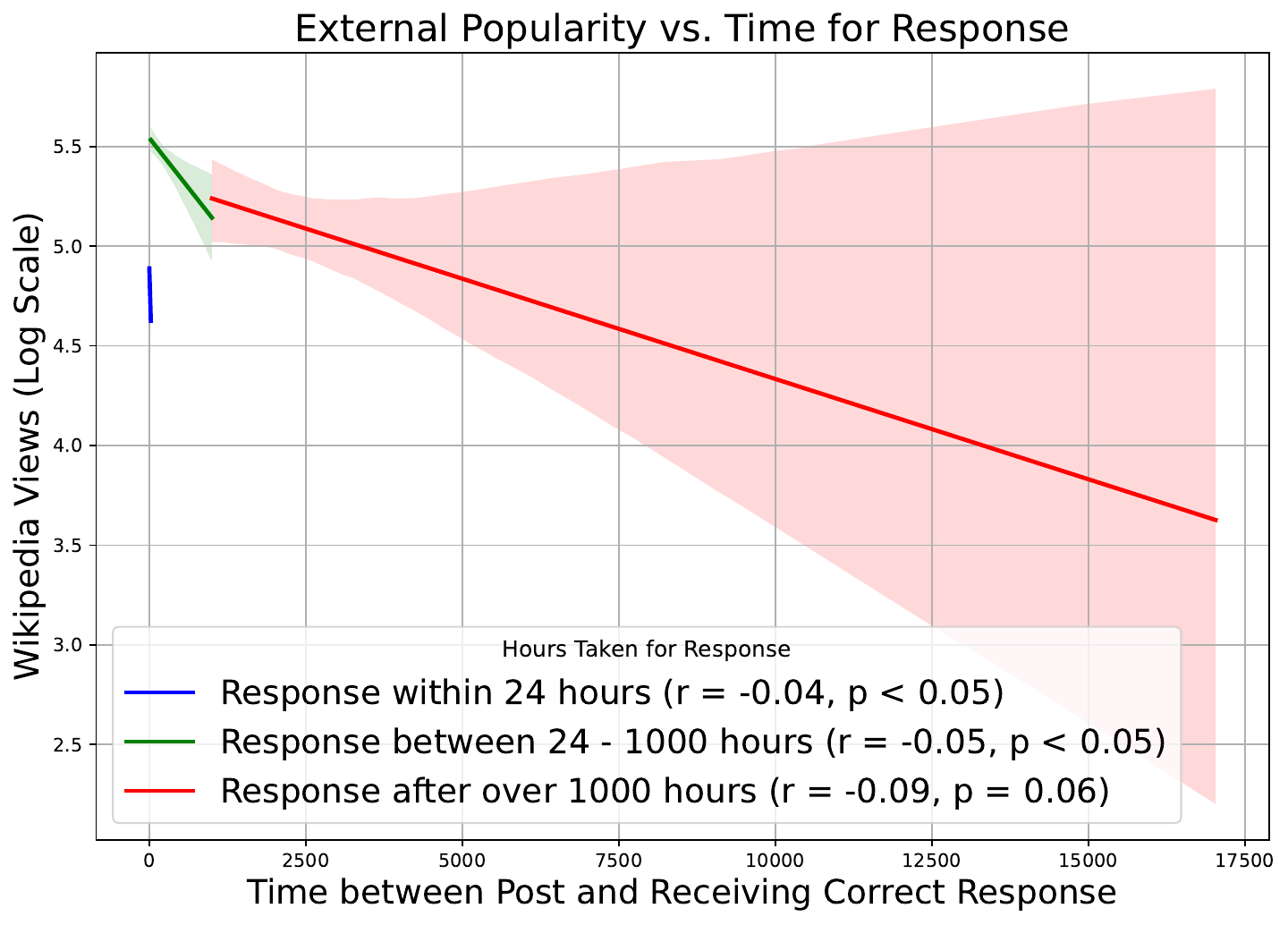}
    \caption{Condensed graph for correlation between Wikipedia page views (popularity) and time taken to receive the correct answer in response.}
    \label{fig:wiki-popularity-versus-response-time-condensed}
\end{figure}

\begin{figure*}
    \centering
    \includegraphics[width=\linewidth]{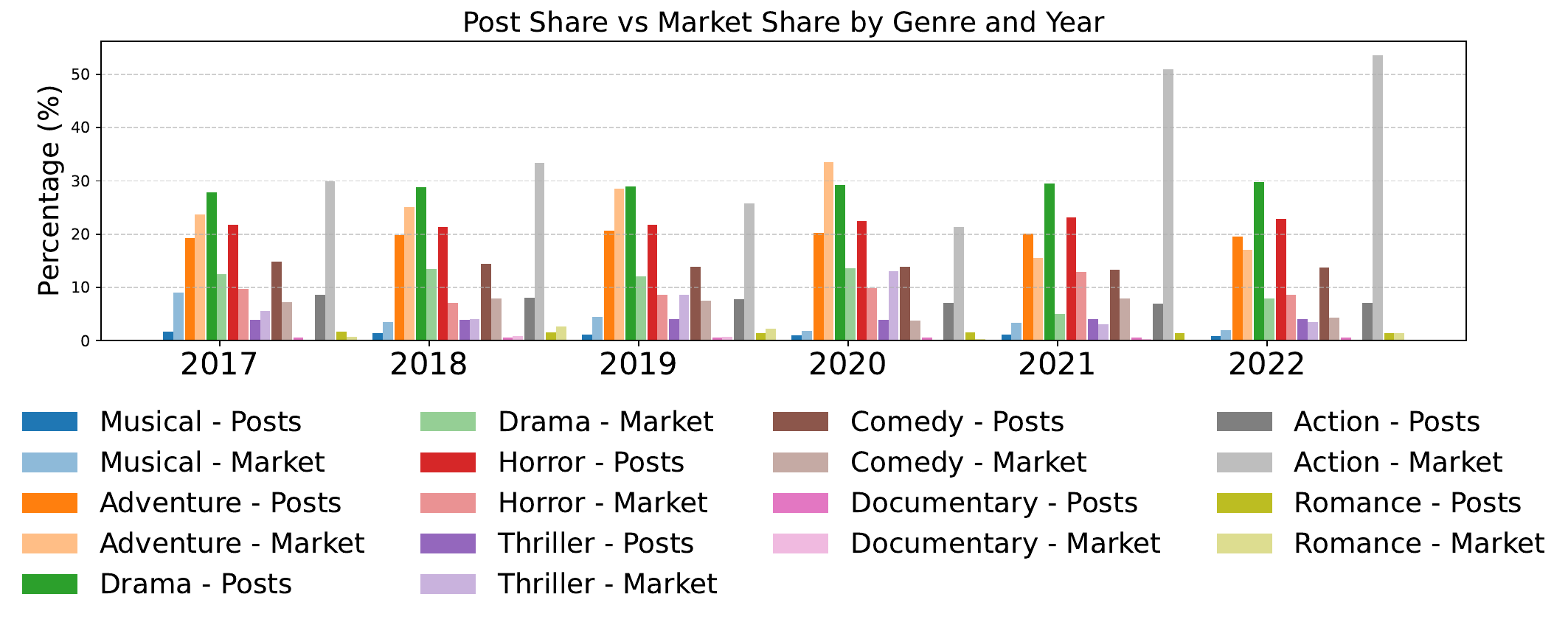}
    \caption{A detailed view comparing genre popularity based on external metrics - in this case, the average revenue earned by each genre within North America - with the corresponding popularity on Reddit ToT search platforms.}
    \label{fig:market-share-vs-posts}
\end{figure*}

\begin{figure}[t]
    \centering
    \includegraphics[width=\linewidth]{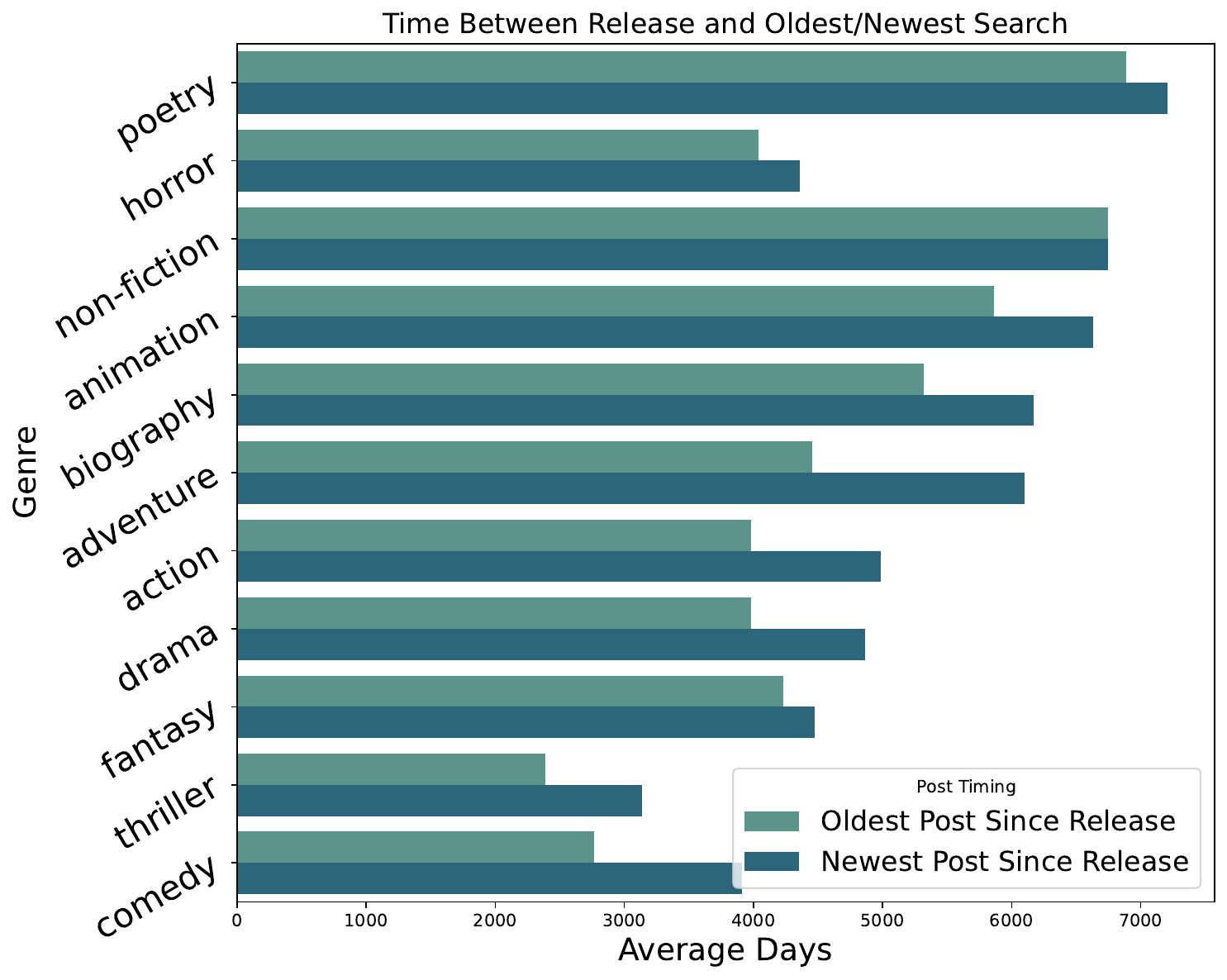}
    \caption{Average number of days since release that the oldest and latest posts are made for each content item, grouped by Genre. The creation date of the Wikipedia page for a given content item is used as the proxy for creation date.}
    \label{fig:post-since-release}
\end{figure}

We include several additional analysis results, complementing the genre-specific and popularity-related analysis, and include them in Figures. \ref{fig:wiki-popularity-versus-reddit-search} to \ref{fig:post-since-release}. 

\section{Details about Implementation}
\label{app:implementation}

\paragraph{Model Training.} Our video-based dataset is split into training and test sets, containing 80,000 and 2500 samples, respectively. For both the recall generation and retrieval task, our model is trained using Low Rank Adaptation \cite{hu2022lora}, with a rank of 64, while keeping the vision model entirely frozen. The training is completed on 4 H100-80 GPUs, and takes 8 hours. We also utilize DeepSpeed Zero and Flash Attention for training. 

\paragraph{Mining Hard Negatives.} Hard negative targets are chosen for each video-recall pair, during the training process for the retrieval task, based on how semantically similar the other recall signals are. For this, we first embed all ground truth recall statements using SBERT \cite{reimers2019sentence}, and compute pairwise cosine similarity. For each sample, we then choose the hard negative recall statement, \(t^-\), by randomly sampling from the top 50 most similar other recall statements. 

Our goal in creating the hard negatives is to ensure that the embedding model learns to distinguish between samples where the memorability signals are the most similar to each other. In the text-based hard negative mining process, capturing direct memorability signals is relatively easy, as we have access to the ground truth recall queries. On the contrary, using multimodal hard negatives is challenging, due to the lack of a precedent method that can embed videos (or find similarities between videos) based on memorability. Thus, we currently focus on using only text-based hard negatives. 

\paragraph{Task Instructions.} Here, we provide the different task instructions used, for the Recall Generation, Prompt Recall Ranking, and Retrieval task: 

\begin{itemize}
    \item \textbf{Instruction for Recall Generation:} \textit{``You are given a detailed description of a video, including the audio transcript of the video, description of each scene in the video, and the text shown in each scene. Your task is to respond with what a person may say, when they are trying to remember the video. Precisely, if a person vaguely remembers the video, and is trying to retrieve a description of the video from their own memory, what are some possible things that they may say? Answer by considering all information about the given video: Audio Transcript: ...., OCR: ....."}
    \item \textbf{Instruction for Prompt Recall Ranking:} \textit{``You will be given multiple images, which are scenes from a video. The images are about some brand, depicting a brand advertisement. There are also several potential descriptions available for the sequence of images, which highlight what is most memorable from the video. Your task is, given 5 candidate descriptions for the images potential descriptions available for the sequence of images, choose the description which is the most fitting. You are required to answer strictly in a JSON format, providing the final answer as follows: {\"answer\": \(<\)Option n\(>\)}"}
    \item \textbf{Instruction for Retrieval:} \textit{``You are given the scenes from an advertisement video, and a detailed description of the video, including description of each scenes in the video, audio transcript of the video, and title of the video. Your task is to respond with what a person may say, when they are trying to remember this advertisement video. Precisely, if a person vaguely remembers the video, and is trying to retrieve a description of the video from their own memory, what are some possible things that they may say? Answer by considering all information about the given video: Audio Transcript: ....., OCR: ......."} 
\end{itemize}

Note that the instruction for the retrieval task is similar to the recall generation task, and it directs the model to learn how to embed the given videos and recall queries to best capture signals of memorability.

\section{Additional Results for Recall Generation}
\label{app:recall_generation}

\begin{table*}[]
    \centering
    \begin{tabular}{ccccccc}
    \toprule
         Model & BLEU & METEOR & ROUGE-1 & ROUGE-2 & ROUGE-L & BERTScore \\
         \midrule
         InternVL-2.5 8B & 0.21 & 0.196 & 0.216 & 0.059 & 0.148 & 0.82 \\
         InternVL 2.5 8B w/ \textsc{ToT2MeM} & \textbf{0.24} & 
         \textbf{0.27} & \textbf{0.28} & \textbf{0.09} & \textbf{0.192} & \textbf{0.85} \\
         \bottomrule
    \end{tabular}
    \caption{Results of fine-tuning another strong baseline using our dataset. This further supports the effectiveness of the dataset, and shows that the gains do not stem from a specific kind of backbone architecture.}
    \label{tab:improved_internvl}
\end{table*}

\begin{table*}[]
    \centering
    \begin{tabular}{ccccccc}
    \toprule
        Model & BLEU & METEOR & ROUGE-1 & ROUGE-2 & ROUGE-L & BERTScore \\
        \midrule
        \textsc{ToT2MeM-Recall} & 0.242 & 0.293 & 0.304 & 0.152 & 0.251 & 0.85 \\
        \textsc{ToT2MeM-Recall} w/ masked OCR, ASR &  0.241 & 0.291 & 0.303 & 0.153 & 0.252 & 0.85 \\
        \bottomrule
    \end{tabular}
    \caption{Results of evaluating our model with test data where proper nouns are masked out from the generated OCR and ASR, to prevent leakage of information. The model performance remains relatively unchanged. }
    \label{tab:masked_ocr_asr}
\end{table*}

Here, we present additional analysis and results for the recall generation task. \\

\noindent \textbf{Fine-tuning other baselines:} As described in \ref{sec:recall}, our presented model \textsc{ToT2MeM-Recall} is a version of Qwen 2.5 VL 7B \cite{bai2025qwen2}, fine-tuned on \textsc{ToT2MeM}. We use this setting to specifically demonstrate the effectiveness of our dataset, showing that without architectural changes, simply fine-tuning a baseline with our dataset can help learn a generalizable memorability signal. As an additional experiment, we also fine-tune InternVL 2.5 8B with \textsc{ToT2MeM}, and find similarly improved performance, as shown in Table \ref{tab:improved_internvl}. \\

\noindent \textbf{Controlling for Proper Noun Leakage in OCR, ASR:} Next, to understand whether the leakage of proper nouns through the automatically generated OCR and ASR provides an unfair advantage to the \textsc{ToT2MeM-Recall} model at inference time, we perform an ablation study. We mask out all of the proper nouns in the ASR and OCR, simply replacing them with an empty string. We avoid using an explicit mask token (e.g., the commonly used [MASK]) to ensure that the model does not get further confused. We use the Spacy library \footnote{https://spacy.io/} to remove proper nouns pertaining to the following: individual people names, names of organizations (e.g., companies, agencies, institutions, etc.), geopolitical entity names (e.g., country, city, state), non-political locations (e.g., mountains, rivers, etc.), creative titles, named historical, cultural, or sports events, named commercial products, and names of manmade facilities or landmarks (e.g., Eiffel Tower, Heathrow Airport, etc.). We manually verify the masking process for 20 samples from the dataset and find that proper nouns are removed from both the transcript and the OCR. We present the ablated results in Table \ref{tab:masked_ocr_asr}. The performance of the model remains virtually unchanged, confirming that the model’s predictions are not driven by name leakage but by broader descriptive signals. It is also worth noting that in some cases, with the proper nouns masked out with an empty string (``"), the audio transcript, in particular, becomes slightly noisy. Our results show that our model is also robust to such variations. In other words, fine-tuning with our dataset equips it with the capability to learn generalizable memorability signals. 

\begin{table*}[h!]
    \centering
    \begin{tabular}{cccc}
    \toprule
         Model & BLEU & ROUGE-1 & BERTScore \\
         \midrule
         \textsc{ToT2MeM-Recall} & 0.242 & 0.304 & 0.85 \\
         \textsc{ToT2MeM-Recall} - OCR & 0.23 & 0.26 & 0.84 \\
         \textsc{ToT2MeM-Recall} - ASR & 0.22 & 0.23 & 0.82 \\
         \bottomrule
    \end{tabular}
    \caption{Comparison of our model with ablated training setups. The ``- OCR" denotes training without the per-scene OCR texts, and ``- ASR" denotes training without the audio transcript for each video.}
    \label{tab:ablation_ocr_asr}
\end{table*}

\noindent \textbf{Ablating OCR and ASR:} We also provide two additional ablation experiments. We ablate both the presence of OCR and the automatically generated transcript in training data, one by one, and show the results on two of the chosen metrics in Table \ref{tab:ablation_ocr_asr}. We find that both the OCR and audio transcripts contribute to the performance of the trained models, with the contribution of ASR being slightly more significant. We hypothesize that there may exist an intuitive reason for this. For videos where the visual content and the semantic meaning (or message) are disparate, the audio transcript provides a bridge between the two. It may also be significantly useful for the model to track temporal changes and relate them to the temporally changing scenes. Particularly, given that we only provide the model with sampled keyframes (limited to 30 scenes per video), the audio transcript becomes the only source of complete information about the video. However, even without either the OCR or the audio transcript, the model is capable of learning some memorability signals from the visual information, leading to improvements over its zero-shot version.


\begin{figure*}
    \centering
    \includegraphics[width=0.9\linewidth]{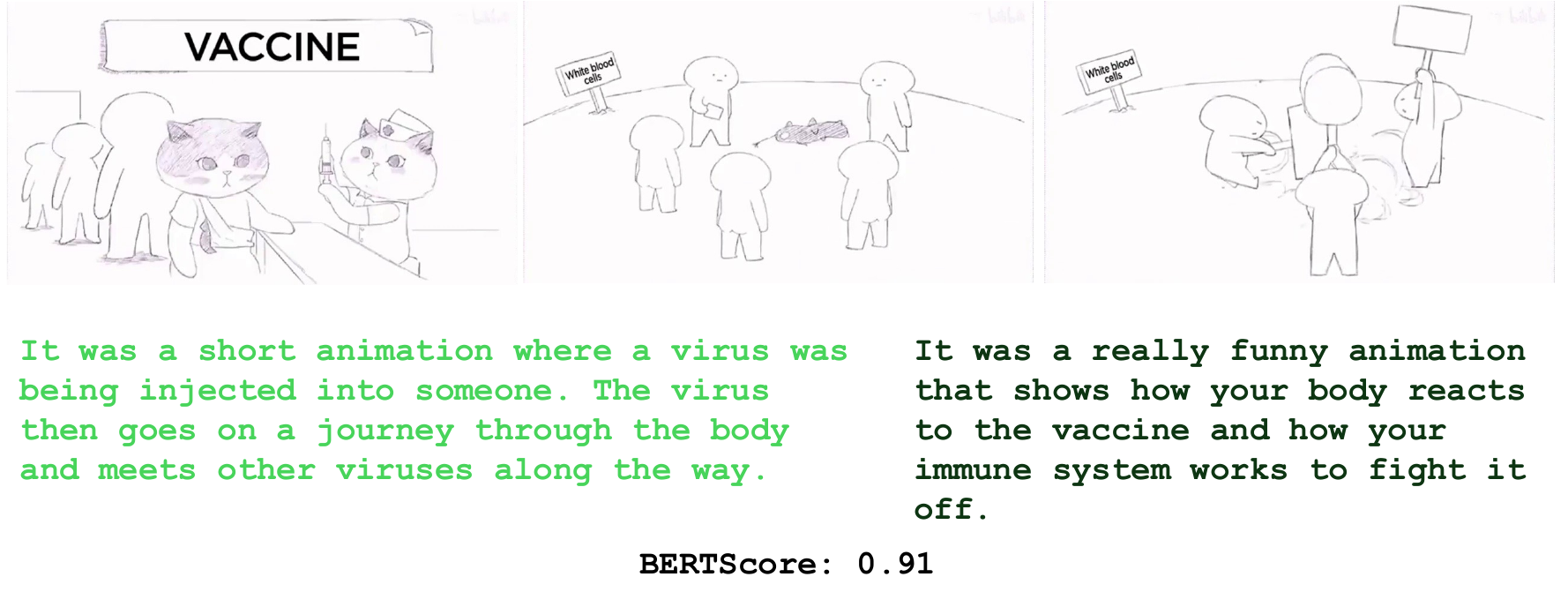}
    \caption{Example 1 of a highly scored recall generated by our model \textsc{ToT2MeM-Recall}. In this example, the text on the \textbf{left shows the prediction}, while the \textbf{text on the right shows the ground truth recall}. The corresponding BLEU Score for the generation is 0.46, and ROUGE-1 score is 0.39.}
    \label{fig:qualitative_recall_1}
\end{figure*}


\begin{figure*}
    \centering
    \includegraphics[width=0.9\linewidth]{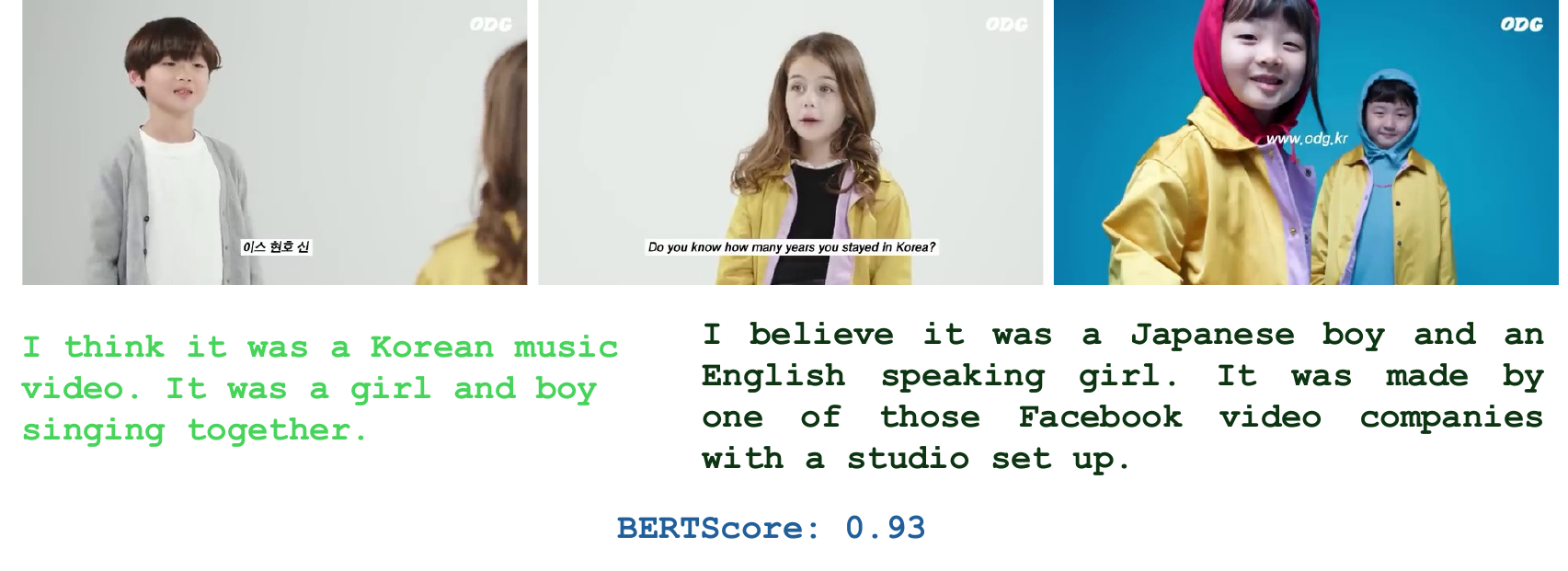}
    \caption{Example 2 of a highly scored recall generated by our model \textsc{ToT2MeM-Recall}. In this example, the text on the \textbf{left shows the prediction}, while the \textbf{text on the right shows the ground truth recall}. The corresponding BLEU Score is 0.41, and ROUGE-1 score is 0.62.}
    \label{fig:qualitative_recall_2}
\end{figure*}


\begin{figure*}
    \centering
    \includegraphics[width=0.9\linewidth]{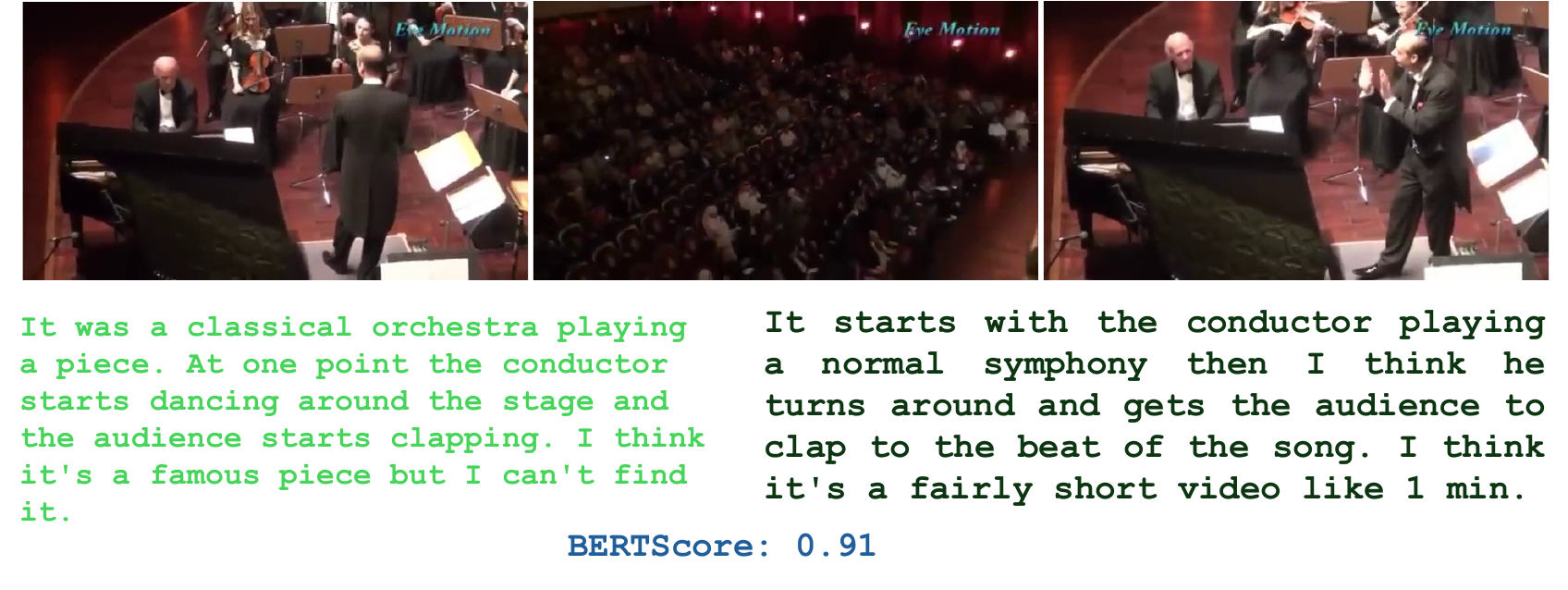}
    \caption{Example 3 of a highly scored recall generated by our model \textsc{ToT2MeM-Recall}. In this example, the text on the \textbf{left shows the prediction}, while the \textbf{text on the right shows the ground truth recall}. The corresponding BLEU Score is 0.61, and the ROUGE-1 score is 0.56.}
    \label{fig:qualitative_recall_3}
\end{figure*}


\begin{figure*}
    \centering
    \includegraphics[width=0.9\linewidth]{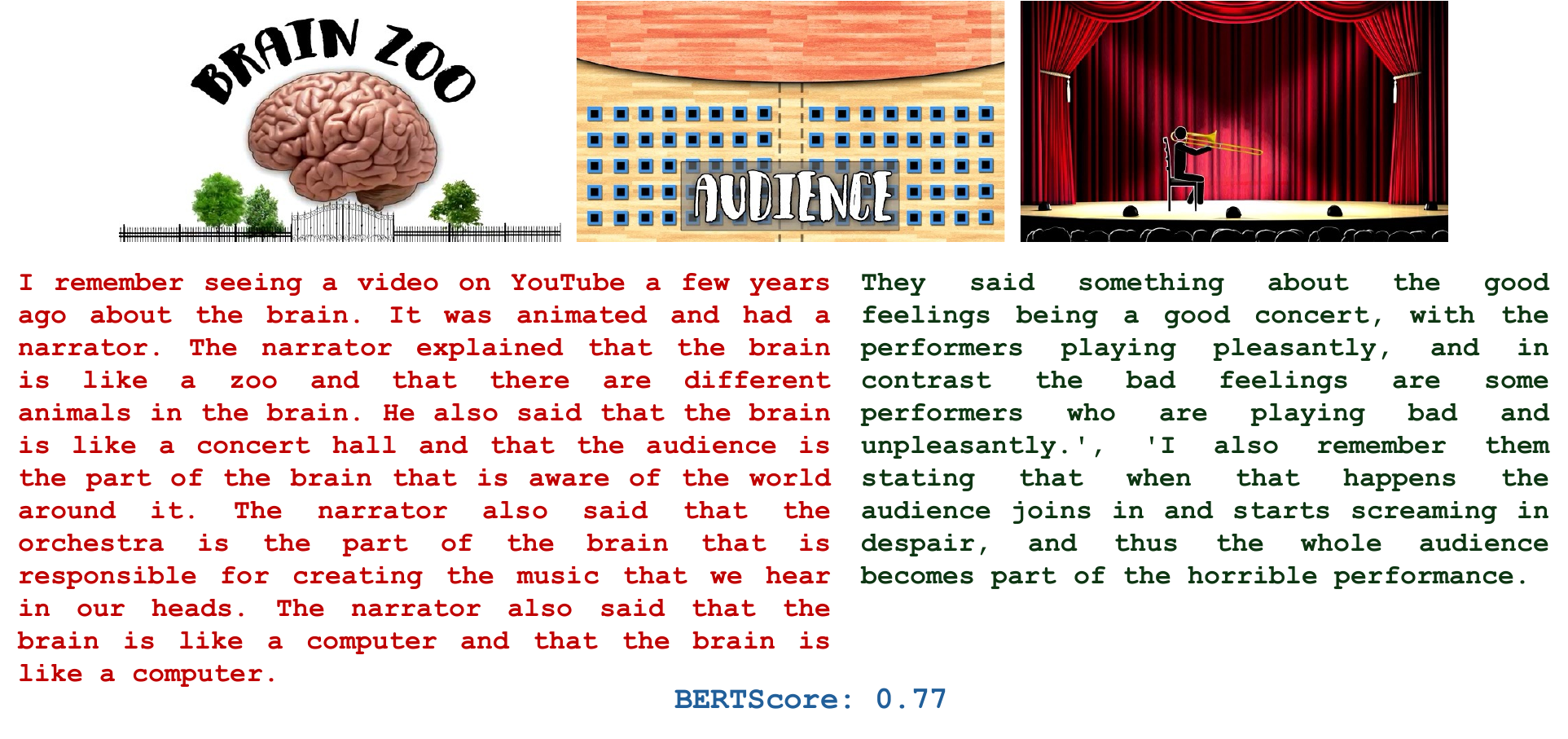}
    \caption{Example 3 of a highly scored recall generated by our model \textsc{ToT2MeM-Recall}. In this example, the text on the \textbf{left shows the prediction}, while the \textbf{text on the right shows the ground truth recall}. The corresponding BLEU Score is 0.10, and ROUGE-1 score is 0.11.}
    \label{fig:qualitative_recall_4}
\end{figure*}


\begin{figure*}
    \centering
    \includegraphics[width=0.9\linewidth]{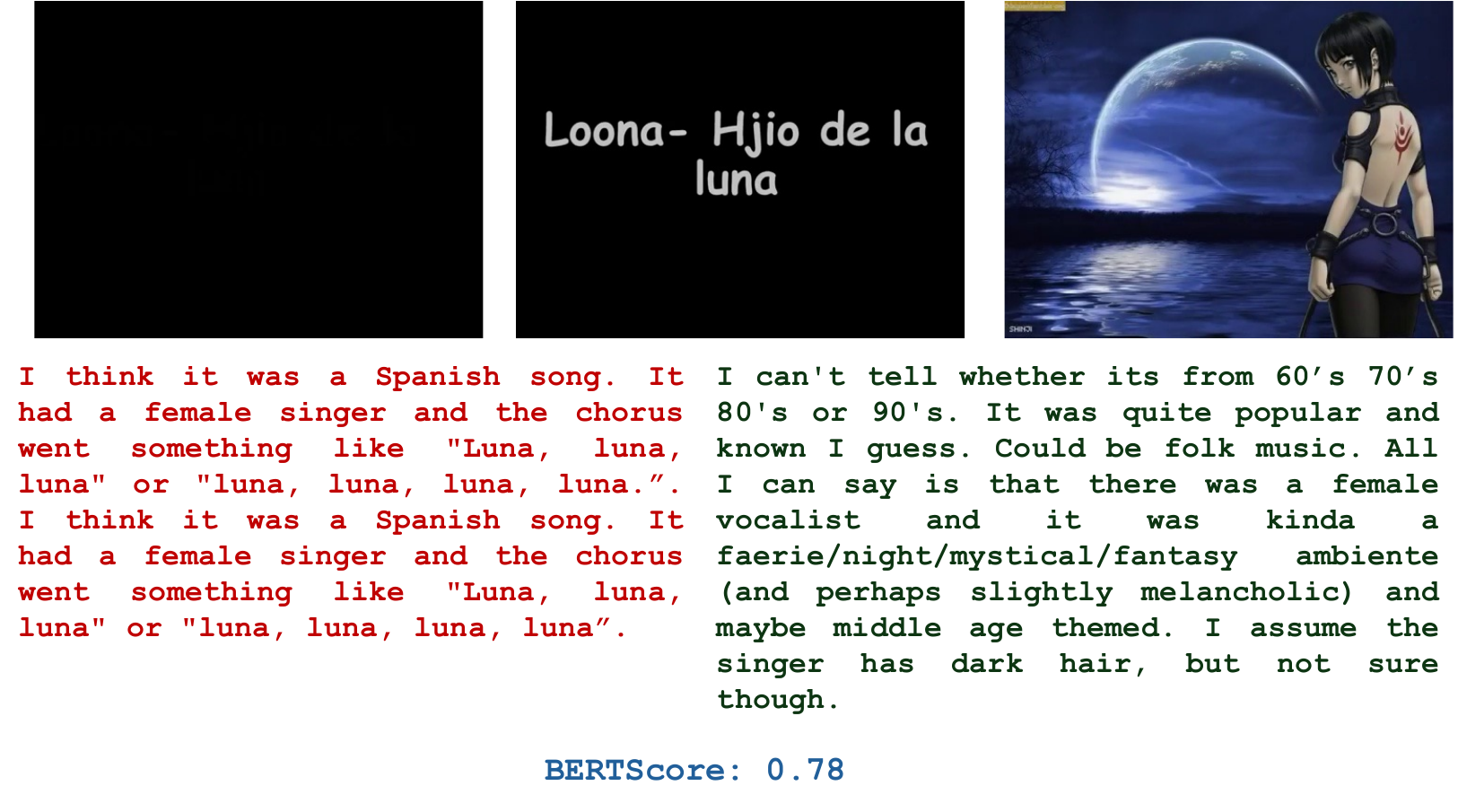}
    \caption{Example 3 of a highly scored recall generated by our model \textsc{ToT2MeM-Recall}. In this example, the text on the \textbf{left shows the prediction}, while the \textbf{text on the right shows the ground truth recall}. The corresponding BLEU Score is 0.10, and ROUGE-1 score is 0.13.}
    \label{fig:qualitative_recall_5}
\end{figure*}


\begin{figure*}
    \centering
    \includegraphics[width=0.9\linewidth]{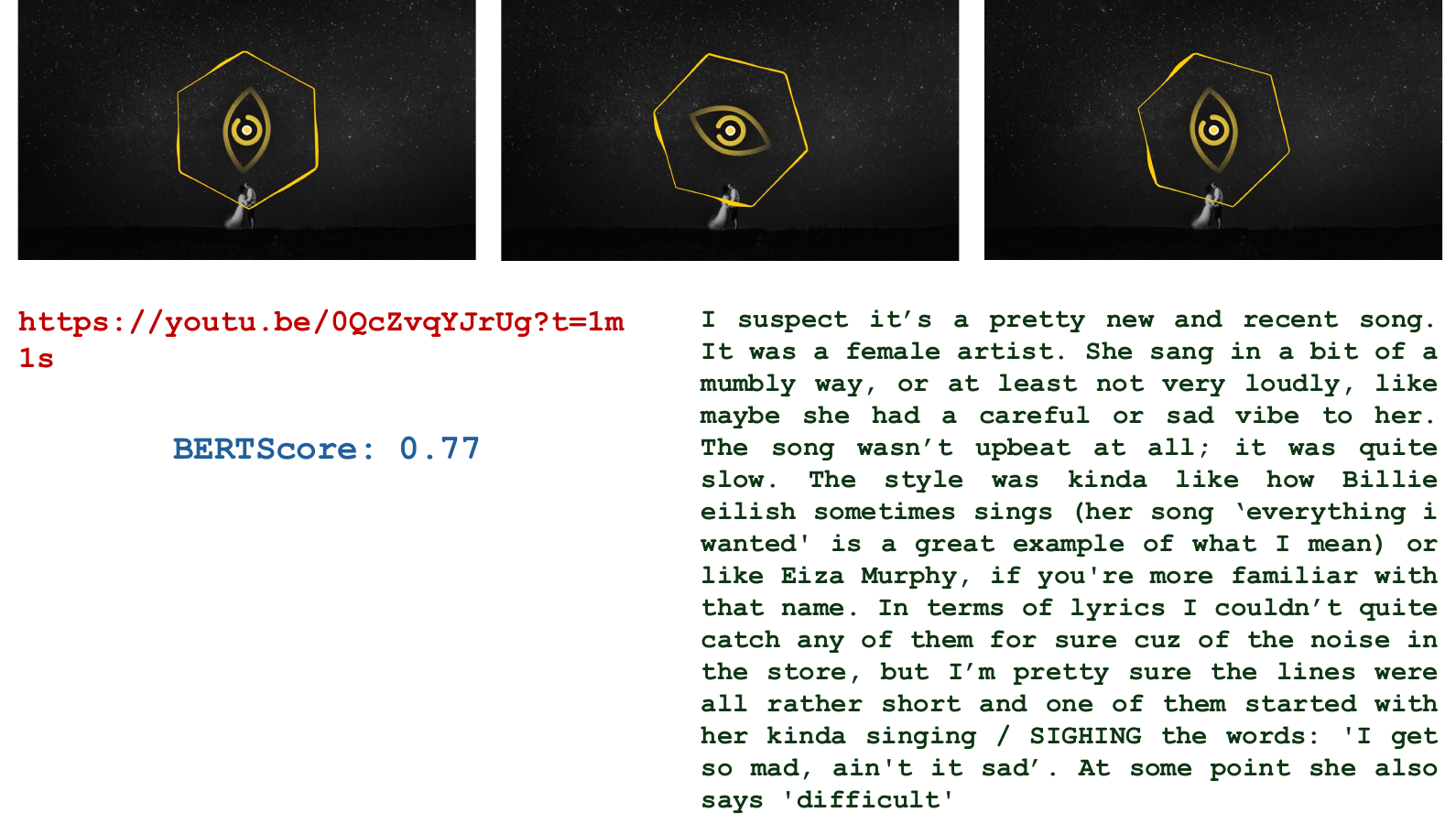}
    \caption{Example 3 of a highly scored recall generated by our model \textsc{ToT2MeM-Recall}. In this example, the text on the \textbf{left shows the prediction}, while the \textbf{text on the right shows the ground truth recall}. The corresponding BLEU Score is 0.00004, and ROUGE-1 is 0.09.}
    \label{fig:qualitative_recall_6}
\end{figure*}


\section{Qualitaive Examples}
\label{app:qualitative_examples}

In this section, we provide qualitative examples of responses generated by the models fine-tuned on our dataset, \textsc{ToT2MeM-Recall} and \textsc{ToT2MeM-Retrieval} for the recall generation and ToT retrieval tasks, respectively. 

\paragraph{Examples for Recall Generation.} Figures \ref{fig:qualitative_recall_1} to \ref{fig:qualitative_recall_6} show the qualitative examples for high-scoring and low-scoring generated recall samples. We use the BERTScore ratings for each generation to retrieve the most highly scored, and the lowest scored examples. BERTScore is chosen to provide the qualitative examples, due to its higher semantic matching capability compared to the other static metrics, like BLEU or ROUGE scores. We show 3 scene examples from each video, and omit directly providing the YouTube video ID or link, to preserve privacy for the video uploaders and channels. 

Figures \ref{fig:qualitative_recall_1} to \ref{fig:qualitative_recall_3} show the top-rated generations by our model. Interestingly enough, in the second example (Fig. \ref{fig:qualitative_recall_2}, the ground truth query refers to the video as a ``Japanese boy and an English speaking girl", while our model predicts that the video was a ``Korean music video". We hypothesize that the OCR is useful in this case, as the subtitles shown in Korean are picked up by the model as relevant memorability signals. Similarly, in the third example \ref{fig:qualitative_recall_3}, our model is capable of capturing the temporal change of actions in the video, as it generates, ``At one point the conductor starts dancing around...". It can also be noted that the model, in alignment with human-generated recall, adds a sentence expressing confusion. 

On the other hand, Figures \ref{fig:qualitative_recall_4} to \ref{fig:qualitative_recall_6} show the generations with some of the lowest scores. We show here 3 different cases of failures:
\begin{itemize}
    \item \textbf{Gap Between Visual Features and Semantics:} We find, as shown in Fig. \ref{fig:qualitative_recall_4}, the generation from our model is highly aligned with the visual content shown in the video, while the ground truth recall query talks more about the semantic content or an abstract summary of the video. This shows that the trained model develops high visual fidelity, and potentially develops a strong visual branch through the training process, different from previous work \cite{harini2025long}, where the model needed explicit textual verbalizations of the scenes to perform adequately well in predicting memorability scores. This also highlights the challenging and nuanced nature of the task, and can inform valuable future work, where the focus can be shifted to building better semantic or abstract representations of videos. This would become especially useful for cases where the visual content may not be well-aligned or entirely coherent with the semantic or underlying message. 
    \item \textbf{Repetitions:} As shown in the second example (Fig. \ref{fig:qualitative_recall_5}, our fine-tuned version of Qwen 2.5 VL, in some cases, falls into the well-known pitfall of repeating tokens. Future work can explore in depth the specific origins of repetition in a memorability-related task \cite{yao2025understanding}, or general-purpose solutions such as employing nucleus sampling \cite{holtzman2020curious}. In fact, we experiment with using nucleus sampling specifically and find that, although the overall performance does not degrade, the issue of tokens being repeated remains.
    \item \textbf{Presence of Links:} As shown in the final error example (Fig. \ref{fig:qualitative_recall_6}), we find a small number of samples where the output is only a link. As previously discussed in Sections \ref{sec:data_collection} and \ref{app:data_collection}, given our automated mechanism of data filtering, for some of the queries, some link content may remain retained. Future work could focus on penalizing the generation of specific kinds of hallucinated text, especially links that do not exist. This could also inform important future work on further exploring what types of hallucinations are possible in the realm of memorability-related tasks. 
\end{itemize}

\begin{figure*}[h!]
    \centering
    \includegraphics[width=0.9\linewidth]{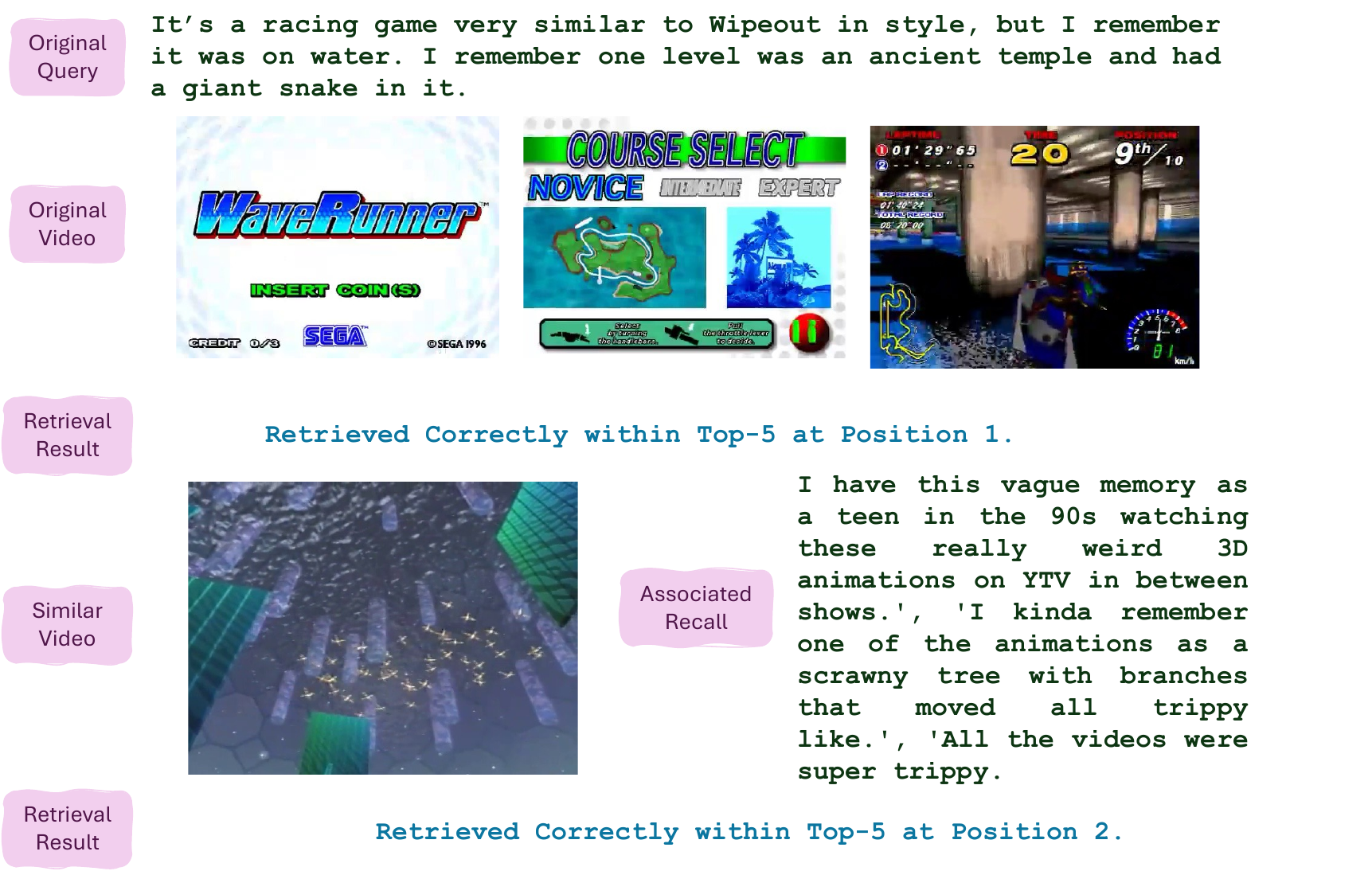}
    \caption{This is an example of a retrieval output, where text-to-video retrieval is performed by our model \textsc{ToT2MeM-Retrieval}. In this case, the correct video is retrieved within the top-5 elements. We show the correct video along with another similar video, which is retrieved within the top-5 items.}
    \label{fig:retrieval_top_5}
\end{figure*}

\begin{figure*}[h!]
    \centering
    \includegraphics[width=0.9\linewidth]{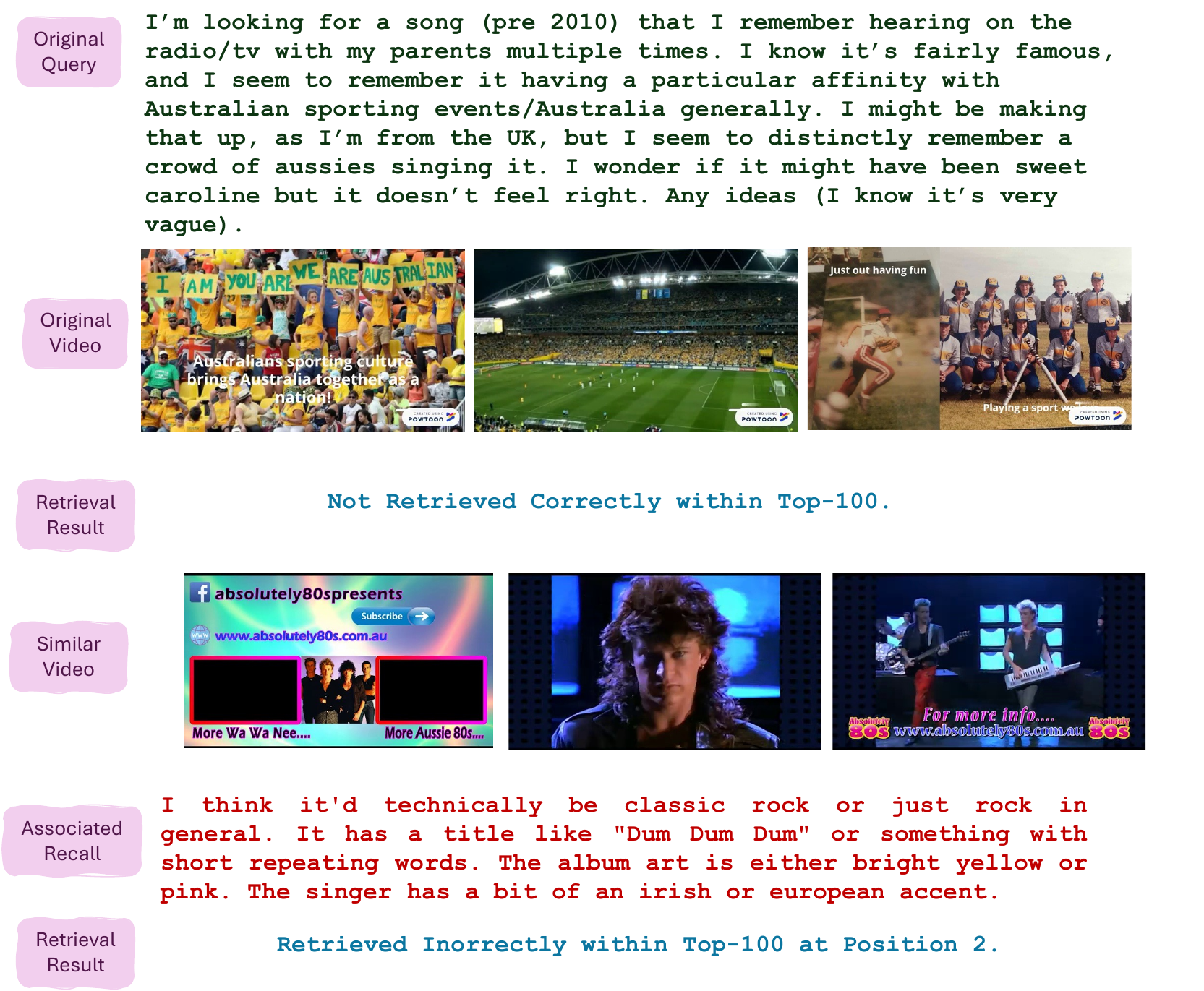}
    \caption{An example of a retrieval output, where the correct video is not retrieved within the top-100 items. In this case, we show the correct video corresponding to the original recall query, which is not retrieved in top-100, and another video, which is incorrectly retrieved within the top-100 items.}
    \label{fig:retrieval_top_100}
\end{figure*}

\paragraph{Examples for Text-to-Video ToT Retrieval.} We further provide similar examples for the multimodal ToT retrieval task. Specifically, we show two types of examples: the first, where the correct video is retrieved within the top-5 items, and the second, where the correct video is not retrieved even within the top-100 items. For each category, we show the original recall query, scenes from the original correct video. For the first category, we also provide examples of one other video which are retrieved in the top-5 items (highly similar to the original video). For the second category, we provide examples of the top-most (incorrect) video. The examples are shown in Figures \ref{fig:retrieval_top_5} and \ref{fig:retrieval_top_100}.

In the correctly retrieved example (Fig. \ref{fig:retrieval_top_5}), the two videos are similar in quality and style. Both are videos of games, and the recall for both gives the description of natural elements. In the incorrectly retrieved example (Fig. \ref{fig:retrieval_top_100}), the common elements between the original and incorrect videos seem to be the content type (song), and mentions of a particular accent, or a particular group of people ("Australian", "European"). Future studies on what is the most optimal information to retain from a recall signal may also be a promising direction.

\section{Additional Details about Human Evaluation}
\label{app:human_evaluation}
As described in Section \ref{sec:recall}, we use a curated experiment involving human data to evaluate the robustness of the recall-generator model, trained on our unsupervised dataset. We collected the data from 163 participants at an academic institution in India. Participation was primarily voluntary, but students were also provided with the option to get optional course credit, or other freebies. The data collection process was reviewed and approved by an IRB within the academic institution. The students were required to sign the IRB consent approval, which was displayed clearly before the data collection. The consent approval information provided details on the kind of data being collected, how the data would be used, and the time commitment required for the study. The participants were provided with the aggregate statistics of the study as a debrief after completing the study. Participants could either take the study in a take-home or in-person format. For take-home participants, similar to the original protocol for the collection of the LAMBDA dataset \cite{harini2025long}, three emails were sent, and if they did not respond, their data was discarded from the experiment. 

The participants were primarily graduate and undergraduate students. The participants are bilingual and speak a variety of languages, including English. The age range of the participants is from 18 to 35 years, and there was again a roughly 30 - 70 \% distribution of female and male participants. We did not record additional gender identification information, although participation from all genders was encouraged. 

Participants were instructed to log into a controlled environment hosted by us and recreate memorable ad scenes using generative models (Stable Diffusion v2.1) \cite{rombach2022high}. The generated images and prompt logs were stored in a submission archive organized by brand and attempt number. The detailed questionnaire for this reconstruction task is included in \S\ref{sec:Scene Reconstruction Questionnaire}.

\subsection{Performance on Prompt Recall Ranking Task}

As seen in \ref{sec:recall}, Table \ref{tab:ranking_evaluation}, there is a significant performance gap between the baseline models InternVL and Qwen 2 VL. From a qualitative analysis of responses, we find that InternVL struggles particularly with producing output in the required format, where it is required to choose between multiple options. On the other hand, Qwen 2 VL excels at providing output in the correct format. The performance shown by our model, \textsc{ToT2MeM-Recall}, becomes particularly noteworthy here. Although it is a model fine-tuned for a completely different task, it is still able to output responses in the correct format, and does so more accurately than the other baselines. We conjecture that the low-rank fine-tuning may have contributed to this, where the model has an improved ability to recognize memorability-related signals, while at the same time being able to respond in accordance with instructions. 

\subsection{Scene Reconstruction Questionnaire}
\label{sec:Scene Reconstruction Questionnaire}
This section contains the additional questions we asked as part of the study, where participants were asked to recreate memorable advertisement scenes using generative models. The reconstructed outputs, prompts, and questionnaire responses were collected for each brand remembered.

\subsubsection{Scene Description and Reconstruction}
\begin{figure}[h]
    \centering
    \includegraphics[width=0.95\linewidth]{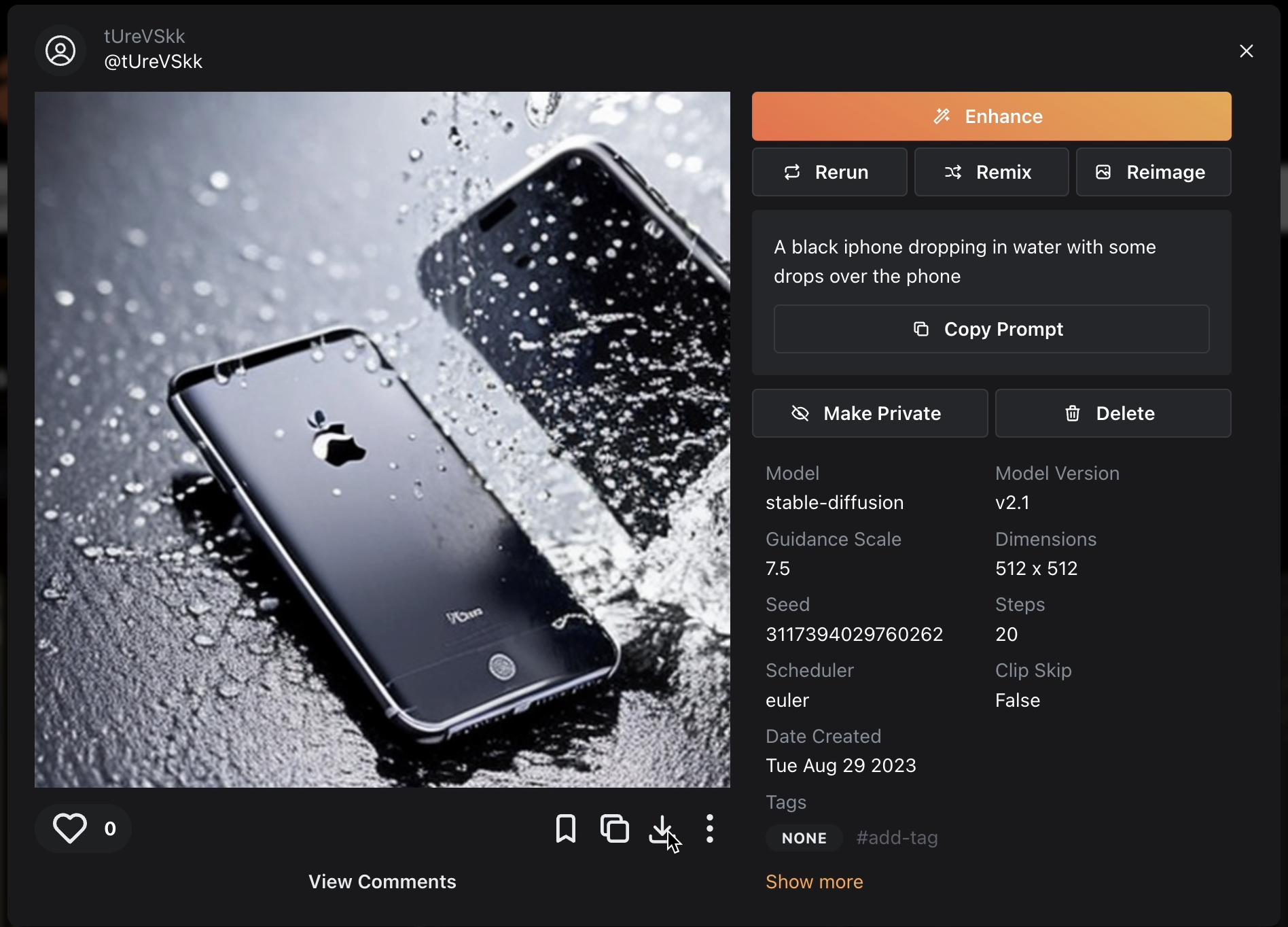}
    \caption{Image generation platform used by students, the prompt can be written here and they can retry variations upto five times.}
    \label{fig:study_screenshot}
\end{figure}

We hosted a website to allow students to simultaneously create scenes, we hosted SD 2.1 on 4 nodes of A100 GPUs. An exemplar scene creation on the platform is shown in figure \ref{fig:study_screenshot}
\begin{enumerate}
    \item For the \{brand\} ad, I remember seeing the following:  
    \begin{itemize}
        \item (Write scene descriptions, feel free to include any scenes, music, characters, emotions, or objects you remember seeing)
    \end{itemize}
    
    \item Please go to the website and recreate a scene that you remember from the \{brand\} ad.  
    \begin{itemize}
        \item Fill your prompts and outputs in \texttt{prompts.txt} and store them in folders \#1, \#2, \#3, \#4, \#5.
        \item You may attempt up to 5 reconstructions per remembered scene.
    \end{itemize}
\end{enumerate}

\subsubsection{Audio Recall (to be filled for each reconstructed brand ad)}
\begin{enumerate}
    \item For the \{brand\} ad(s), what type of audio did you hear? (Select all that apply)
    \begin{itemize}
        \item[a.] Narration
        \item[b.] Background Music
        \item[c.] Silent
        \item[d.] Don’t Remember
    \end{itemize}
\end{enumerate}

\subsubsection{Product Familiarity (to be filled after reconstruction)}
\begin{enumerate}
    \item How many times in the last 1 year have you used the product shown in the \{brand\} ad(s)?
    \begin{itemize}
        \item[a.] 0
        \item[b.] 1--10
        \item[c.] 10+
    \end{itemize}
    
    \item Have you ever used \{brand\} before?
    \begin{itemize}
        \item[a.] Yes
        \item[b.] No
    \end{itemize}
\end{enumerate}

\end{document}